\documentclass[manuscript,screen]{acmart}

\usepackage{subfig}
\usepackage{siunitx}
\usepackage{xcolor, colortbl}
\usepackage{makecell}
\usepackage{tabularx,booktabs}

\AtBeginDocument{%
  }

\setcopyright{acmlicensed}
\copyrightyear{2024}
\acmYear{2024}
\acmDOI{XXXXXXX.XXXXXXX}

\setcopyright{acmlicensed}
\copyrightyear{2018}
\acmYear{2018}
\acmDOI{XXXXXXX.XXXXXXX}

\acmJournal{JACM}
\acmVolume{37}
\acmNumber{4}
\acmArticle{111}
\acmMonth{8}

\begin{document}

\title[Human-in-the-loop Fairness: Integrating Stakeholder Feedback]{Human-in-the-loop Fairness: Integrating Stakeholder Feedback to Incorporate Fairness Perspectives in Responsible AI}

\author{Evdoxia Taka}
\email{evdoxia.taka@glasgow.ac.uk}
\affiliation{%
  \institution{University of Glasgow}
  % \streetaddress{P.O. Box 1212}
  % \city{Glasgow}
  \country{UK}
}

\author{Yuri Nakao}
\email{nakao.yuri@fujitsu.com}
\affiliation{%
  \institution{Fujitsu Limited}
  \country{Japan}
}

\author{Ryosuke Sonoda}
\email{sonoda.ryosuke@fujitsu.com}
\affiliation{%
  \institution{Fujitsu Limited}
  \country{Japan}
}

\author{Takuya Yokota}
\email{yokota-takuya@fujitsu.com}
\affiliation{%
  \institution{Fujitsu Limited}
  \country{Japan}
}

\author{Lin Luo}
\email{l.luo.1@research.gla.ac.uk}
\affiliation{%
  \institution{University of Glasgow}
  \country{UK}
}

\author{Simone Stumpf}
\email{simone.stumpf@glasgow.ac.uk}
\affiliation{%
  \institution{University of Glasgow}
  \country{UK}
}

\renewcommand{\shortauthors}{Taka et al.}

\begin{abstract}
Fairness is a growing concern for high-risk decision-making using Artificial Intelligence (AI) but ensuring it through purely technical means is challenging: there is no universally accepted fairness measure, fairness is context-dependent, and there might be conflicting perspectives on what is considered fair. Thus, involving stakeholders, often  without a background in AI or fairness, is a promising avenue. Research to directly involve stakeholders is in its infancy, and many questions remain on how to support stakeholders to feedback on fairness, and how this feedback can be integrated into AI models. Our work follows an approach where stakeholders can give feedback on specific decision instances and their outcomes with respect to their fairness, and then to retrain an AI model. In order to investigate this approach, we conducted two studies of a complex AI model for credit rating used in loan applications. In study 1, we collected feedback from 58 lay users on loan application decisions, and conducted offline experiments to investigate the effects on accuracy and fairness metrics. In study 2, we deepened this investigation by showing 66 participants the results of their feedback with respect to fairness, and then conducted further offline analyses. Our work contributes two datasets and associated code frameworks to bootstrap further research, highlights the opportunities and challenges of employing lay user feedback for improving AI fairness, and discusses practical implications for developing AI applications that more closely reflect stakeholder views about fairness.
\end{abstract}




\received{30 September 2024}
\received[revised]{12 March 2009}
\received[accepted]{5 June 2009}

\maketitle

\section{Introduction}
The use of AI in high-risk decision-making has risen dramatically and fairness is increasingly recognised as an important concern when deploying AI solutions~\cite{mehrabi2022,Madaio2020Co-Designing, kirkpatrick_battling_2016}. To date, a number of tools have been developed to help AI experts assess and mitigate bias and unfairness \cite{bellamy_ai_2019, ahn_fairsight:_2019, cabrera_fairvis_2019} alongside approaches to remove bias from AI models \cite{mehrabi2022}. While great strides have been made in this direction, there are still a number of major issues to overcome: there is no universally accepted fairness metric, views of AI experts and other stakeholders on fairness might diverge \cite{jakesch_how_2022}, fairness metrics are context-dependent on their application and domain \cite{Holstein2019Improving}, there are currently over 20 fairness metrics in existence, which often conflict \cite{verma2018fairness}, and there might be other, non-protected attributes that are considered important in decision-making \cite{kasinidou_i_2021}. In short, fairness is a human value and therefore purely technical, quantitative means of evaluating fairness have been increasingly questioned as to their appropriateness ~\cite{Binns2018Its,DeCremer2024AI}.

Obtaining input from stakeholders is one way to tackle these issues \cite{Holstein2019Improving, lee2020human} and could remove the need for AI experts to make decisions on which fairness metric to select and the attributes to apply them on, thus ensuring that diverse perspectives are taken into account~\cite{Nakao2023Stakeholder}. However, approaches of seeking and integrating lay users' input about AI fairness are in their infancy. There are several different approaches for obtaining stakeholder feedback \cite{SAXENA_fairness_fare, Srivastava2019Mathematical, cheng2021soliciting, nakao2022}. Our work builds on allowing stakeholders to inspect and provide feedback on decision instances and their predicted outcomes \cite{nakao2022}. We also investigate whether it is possible to identify what fairness metrics stakeholders choose and which features are involved, leveraging work on fitting metrics to user feedback \cite{SAXENA_fairness_fare, Srivastava2019Mathematical}. Our aim is to evaluate a number of ways to integrate stakeholder feedback and the effects on fairness into an overall global model and also into their own individual, personalized models. 

To do so, we ran two studies. In study 1, we collected feedback from 58 lay users on the fairness of loan application predictions. We investigated different approaches of integrating their feedback by offline experiments, by building a global model retrained on all user feedback as well as retraining personalized models for each participant. We then conducted a set of analyses to rigorously investigate the impact of this feedback on individual and group fairness, using a large set of fairness metrics. Building on the above, we conducted a second study. In study 2, we investigated how lay users provide feedback on the fairness of the model in an interactive machine learning (IML) setting. In this study, participants received information about how their feedback influenced the fairness of the AI model while they interacted with the predictions. This study extends our findings from study 1. Our research contributes the following:
\begin{itemize}
    \item two open datasets of lay user feedback on the fairness of loan applications which can be used to stimulate further analyses and research;
    \item code and evaluation frameworks suitable for replicating our work and building new approaches for integrating stakeholder feedback;
    \item empirical results of integrating lay user feedback in AI models to address fairness;
    \item baseline approaches to integrate user fairness feedback in XGBoost and other complex models;
    \item a better understanding of the opportunities and challenges of taking lay user feedback on AI model fairness into consideration;
    \item a set of practical implications for human-in-the-loop fairness approaches that better reflect stakeholder views about fairness.  
\end{itemize}

The structure of our paper is as follows. We first present related work, covering common approaches for measuring and mitigating fairness as well as human-centered research into fairness and involving stakeholders. We then describe our first user study, using the feedback for offline retraining of models. We then describe a second study, in which we retrained personalized models online. Finally, we discuss the limitations of this work, the implications of our results across all studies, and the possibilities for future research.

\section{Related Work}

\subsection{Measuring Fairness and Bias Mitigation}
There are many definitions of fairness, aligned with justice, for example, \emph{interactional fairness}  (treating people with respect)  ~\cite{bies1987interactional,ColquittOrganizationalJustice}, \emph{procedural fairness} (fairness is made in a logical, correct manner) ~\cite{Robert2020Designing,ColquittOrganizationalJustice} and \emph{outcome fairness} (outcomes are fairly allocated or distributed) ~\cite{adams1963towards,rawls2001justice,ColquittOrganizationalJustice}. There is a growing body of work exploring more fairness perspectives of stakeholders \cite{jakesch_how_2022, kasinidou_i_2021, cheng2021soliciting, Holstein2019Improving}. Most commonly in AI, fairness has been considered from a quantitative perspective, looking at outcome fairness \cite{verma2018fairness, mehrabi2022, bellamy_ai_2019}. Outcome fairness can be categorized in different ways e.g., \emph{group fairness}~\cite{hardt2016equality,dwork2012fairness,berk2021fairness}, \emph{counterfactual fairness}~\cite{kusner2017counterfactual}, \emph{subgroup/intersectional fairness}~\cite{kearns2018Preventing,yang2020fairness,foulds2020intersectional,kobayashi2022One}, and \emph{individual fairness} \cite{zemel2013learning,dwork2012fairness,speicher2018unified}.  For each type of categorization, there are multiple metrics to evaluate fairness in AI, often measured on ``ground truth'' data. 

Group fairness often considers outcomes across a restricted set of \emph{protected} demographic attributes, such as gender, race/ethnicity, disability, age, and marital status as enshrined in law. To measure group fairness, equalized odds~\cite{hardt2016equality}, equal opportunity~\cite{hardt2016equality}, statistical parity~\cite{dwork2012fairness}, and conditional statistical parity~\cite{berk2021fairness} are often employed. Similar metrics exist for other kinds of group fairness, for example, counterfactual fairness, which assesses how much the outcome an individual receives would change if their attributes are taken into account~\cite{kusner2017counterfactual}, and subgroup or intersectional fairness, which aims to ensure fairness among groups where multiple attributes are taken into account~\cite{kearns2018Preventing,foulds2020intersectional}. For individual fairness, which aims to ensure that individuals who are similar receive similar outcomes, several metrics such as Consistency~\cite{zemel2013learning}, L-Lipschitz Continuity~\cite{dwork2012fairness}, and Theil index~\cite{speicher2018unified} have been implemented. However, many of these metrics are based on measuring similarity between individuals which is contentious~\cite{cheng2021soliciting}.

Various bias mitigation approaches have emerged to tackle unfairness \cite{hort2022bia}, using pre-, in-, and post-processing interventions in the AI model development pipeline. \emph{Pre-processing mitigation} involves modifying the raw data to eliminate or reduce potential biases  \cite{Chakraborty2020Fairway}, such as relabeling data \cite{kamiran2009classifying,kamiran2012data}, perturbing data attributes \cite{feldman2015certifying,wang2019repairing}, reweighing samples' importance \cite{calders2009building,krasanakis2018adaptive}, or altering data distribution by sampling \cite{kamiran2012data}. \emph{In-processing mitigation} includes optimizing hyper-parameters \cite{Chakraborty2020Fairway}, regularizing to modify loss functions \cite{zhang2022longitudinal,Kamishima2012Fairness}, and active learning-based fairness mitigation \cite{noriega2019active,anahideh2022fair} while training the model. \emph{Post-processing mitigation} addresses bias in already trained models, e.g., by modifying the decision boundary \cite{iosifidis2019fae} or prediction results \cite{kamiran2018exploiting}, or interactive model selection after training \cite{Ashktorab2023}. Many toolkits have been developed to facilitate evaluating and mitigating bias, for example, AI Fairness 360 \cite{bellamy_ai_2019}, Fairlearn \cite{weerts2023fairlearn}, and the What-If Tool \cite{wexler2019if}. However, bias mitigation approaches usually target a specific fairness metric and a subset of protected attributes. Given the context-dependent nature of fairness metrics, these mitigation approaches need to be applied cautiously \cite{lee2021landscape}. It has been argued that mitigating bias automatically cannot be called fair as this inevitably defines fairness quite narrowly. Thus, questions have arisen of who should define what fairness is and how, placing greater power in the hands of non-technical stakeholders ~\cite{Holstein2019Improving,Sloane2022Participation}.  

\subsection{Human-Centered Fairness}

 There are numerous challenges in approaching fairness purely quantitatively: stakeholders often differ in their fairness definitions~\cite{RN54}, there is a multitude of conceptions of fairness, no ``one-size-fits-all'' fairness metric, and fairness metrics are sometimes incompatible~\cite{Binns2018Fairness}. Thus, metric selection for specific contexts is challenging~\cite{Shneiderman2020Human,Srivastava2019Mathematical,corbett2017algorithmic}. Obtaining input from stakeholders could help to solve this challenge \cite{Holstein2019Improving, lee2020human}, removing the need for AI experts to make decisions on which fairness metric to select and the attributes to apply them on.

Some existing work has created user interface (UI) systems that allow the fairness of AI models to be \emph{inspected}. Among them are tools designed by both academia~\cite{cabrera_fairvis_2019,ahn_fairsight:_2019} and industry~\cite{bellamy_ai_2019,Wexler2020WhatIfTool,bird2020fairlearn} to be used by AI experts and data scientists.  
% (FairVis, AIF360, WhatIfTool, FairSight, RAIDashboard)
These tools require knowledge of AI and fairness metrics and are not supposed to be used by a wide range of stakeholders, including domain experts and end-users. Other UIs are aimed at stakeholders without a background in AI. For example, \citet{cheng2021soliciting} developed a UI to help parents and social workers evaluate the fairness of AI models for child maltreatment cases. In the UI, stakeholders explore decisions made by an AI about cases, and then pick fairness metrics which are explained to them. Similarly, \citet{10.1145/3579601} developed a visualization of fairness metrics calculated on a model for graduate school admissions.  
Another system, DiscriLens, used an UI with visualizations to present a set of potentially discriminatory decisions~\cite{Wang2021Visual}. In our previous work, we designed UIs for loan applications with domain experts such as loan officers ~\cite{RN54} and non-expert decision subjects i.e. potential loan applicants ~\cite{nakao2022}, to assess the fairness of an AI model. 

Work on \emph{incorporating} non-expert stakeholders' feedback in AI fairness is nascent. There are several different approaches. One approach is by asking stakeholders to give feedback on preferred outcomes of decision scenarios instances and then ``retrofitting'' the appropriate fairness metric against the feedback \cite{SAXENA_fairness_fare, 10.1145/3375627.3375862}. A drawback of this approach is that the scenarios and the responses must be carefully chosen and it is difficult to see how stakeholders are able to distinguish between many but subtly different fairness metrics. Another approach is to show a series of decision outcomes from two AI models against the ground truth and ask stakeholders to pick the preferred model \cite{Srivastava2019Mathematical}. The feedback is then used to determine the fairness metric that fits the responses best. Similarly, this requires careful choice of models to expose to stakeholders. 

A different approach is to allow stakeholders to compare decision instances and their predicted outcomes, as well as providing information on model fairness employing different metrics~\cite{cheng2021soliciting}. However, this relies on explaining fairness metrics to stakeholders and their nuanced implications. Last, our previous work \cite{nakao2022} has explored how to support lay users in identifying ``fair'' and ``unfair'' decision instances and their outcomes, and using this feedback to improve fairness, inspired by work in Explanatory Debugging to steer AI models \cite{Kulesza2015Principles}. However, many questions remain how stakeholder feedback can be integrated into AI models, and whether this makes the entire AI model fairer. In our previous work \cite{nakao2022}, we have already shown that allowing lay users to provide feedback on model predictions and making feature weight adjustments can increase fairness measured through \emph{one} fairness metric, for \emph{one} specific non-protected feature of a retrained model. We extend this work in this paper in four main aspects. First, the dataset is much larger and more complex than the one used in the previous work. Second, the previous dataset is not publicly available so our work cannot be replicated. Third, the XGBoost model we use in this work is more complex than the previous logistic regression model, mirroring advances in AI. Last, the previous work only used one group fairness metric and one attribute while this work evaluates fairness against seven different fairness metrics spanning group and individual fairness and several different protected features, thus investigating more deeply how user feedback affects the fairness of a model based on multiple metrics.

Our approach is closely tied to efforts to involve end-users in directly shaping an AI model, to make it more ``correct'' and thus also fairer, similar to research on integrating lay users' knowledge to adjust algorithmic outcomes \cite{10.1145/3359284}. Involving end-users in adjusting the decision-making of an AI model has been explored in the field of \emph{active learning (AL)} and \emph{interactive machine learning (IML)}. In AL~\cite{settles2009active}, humans, as oracles,  provide annotations on \textit{system-selected examples}.
In contrast, IML \cite{fails2003interactive}, which also treats humans as oracles, is a paradigm involving diverse stakeholders in tightly coupled and incremental AI development loops \cite{RN168,amershi2014power,RN116, ramos2020interactive} but using \textit{user-generated or user-selected} examples. A benefit of IML is rapid model update \cite{maadi2021review,gillies2016human,RN147}, while allowing seamless human-AI interactions for data production and selection, and model evaluation and refinement \cite{sakata2019crownn,huang2019combining, kapoor2010interactive,maadi2021review}. Hence, our work treats stakeholders as ``fairness oracles''.

\section{User Study 1}
We carried out a user study to elicit user feedback on the fairness of loan application decisions produced by an AI model. The collected feedback was then integrated in a variety of ways in offline experiments into a global model and per-participant personalized models.

\subsection{Method}
\subsubsection{The Dataset and AI Model}

% The AI system in \citet{nakao2022} constituted a logistic regression model trained on a loan applications dataset provided by a  financial partner to predict acceptance or rejection of applications. 
The code for preparing the data and training the model can be found at
\url{https://github.com/evdoxiataka/effi_user_study1}. We used an open-source loan application dataset, the Home Credit dataset \cite{homeCreditDataset}. The training instances noted clients with payment difficulty, and in our user study we used this as the basis for predicting acceptance or rejection of clients' applications. We removed 49 features with more than $25\%$ missing values and another $22$ features where the data dictionary did not list clear definitions or provenance from the original dataset. From the final $49$ features that were used for the training, $3$ were protected attributes (Gender, Marital Status, Age). Missing values were appropriately imputed. Normalization of amount features with very large numerical values, and appropriate encoding of categorical variables were applied. Some labels and values were transformed before displaying the instances to lay user participants to ease understandability.

%The dataset consists of $307,511$ instances of training set including the Target variable ($24,825$ rejected and $282,686$ accepted applications) and $48,744$ instances of test set without the Target variable.

We chose a XGBoost decision tree classifier \cite{XGBoost, Chen2016} to develop the model since it has been often applied to this dataset and has achieved good results. This type of classifier offers also a natural way of integrating and interpreting user feedback. For example, the model's parameter of \emph{feature weights} can be manipulated to calibrate the feature importance set in the model's decisions. %Class weights were used at the time of training to deal with the class imbalance. % We extracted the feature weights from the trained XGBoost classifier. 

The dataset consists of two sets: Set 1, a training set, and Set 2, a test set. Because it was originally created for a competition, Set 1 includes the Target variable, Set 2 does not. We trained the model using $70,000$ instances from Set 1, undersampling to deal with the class imbalance. Similarly, we created a hold-out set of $30,000$ instances from Set 1 to test the model(s). We used the first $1,000$ instances of Set 2 to predict application outcomes using the trained model. Participants were then provided with these predictions, along with extracted feature weights, to collect their feedback on the fairness of the model's decision. For our offline experiments, we retrained models on the subset of Set 1 used initially to train the model, extended by the instances with feedback from the user study i.e. the subset of Set 2. We explored various ways of integrating the feedback instances into the training set (see Section 3.1.6 Offline Experiments) and evaluated the retrained models against the hold-out dataset from S1.

\subsubsection{The User Interface}
As outlined in Related Work, there have been a number of UIs designed, implemented and evaluated that allow users to feed back on class labels \cite{RN177,RN116} and AI fairness \cite{wexler2019if, ahn_fairsight:_2019, cheng2021soliciting, RN54}. We implemented a UI to present the details of loan applications, to explain how the AI system made decisions, and to enable users to provide feedback, closely based on components put forward and validated by our previous work \cite{nakao2022, RN54}. 

%We now provide a very brief overview of the UI's main functionality to provide user feedback; more detailed information and a screenshot of the UI can be found in Appendix \ref{sec:App_UI}. 

\begin{figure}[ht]
\centering
\subfloat[]{%
\includegraphics[width=0.6\textwidth]{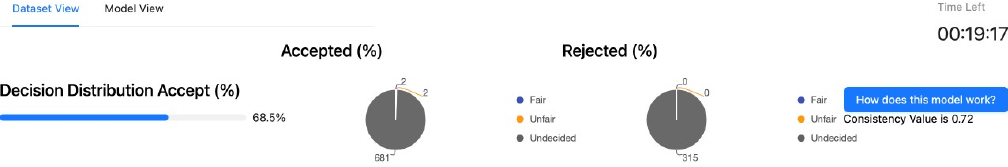}\label{fig:fig1a}
}\\
\subfloat[]{%
    \includegraphics[width=0.6\textwidth]{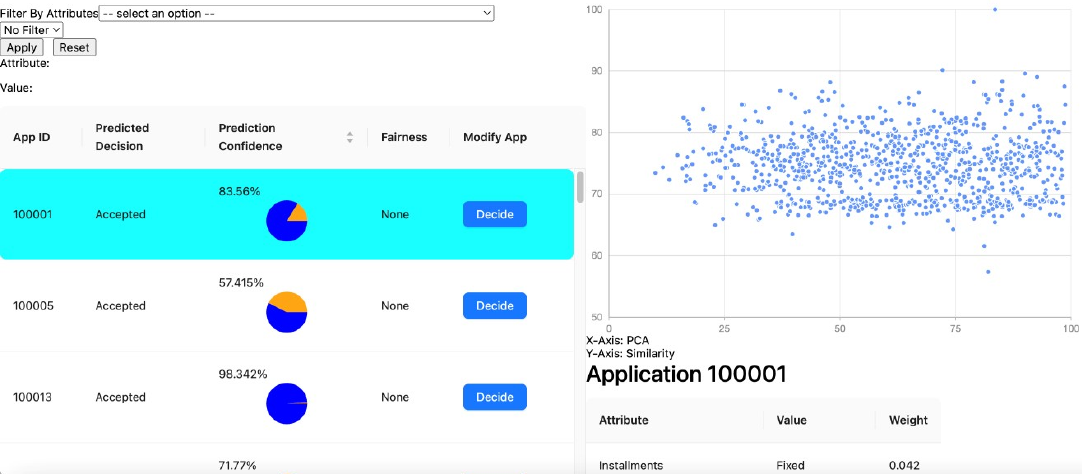}\label{fig:fig1b}}\quad
\subfloat[]{%
    \includegraphics[width=0.37\textwidth]{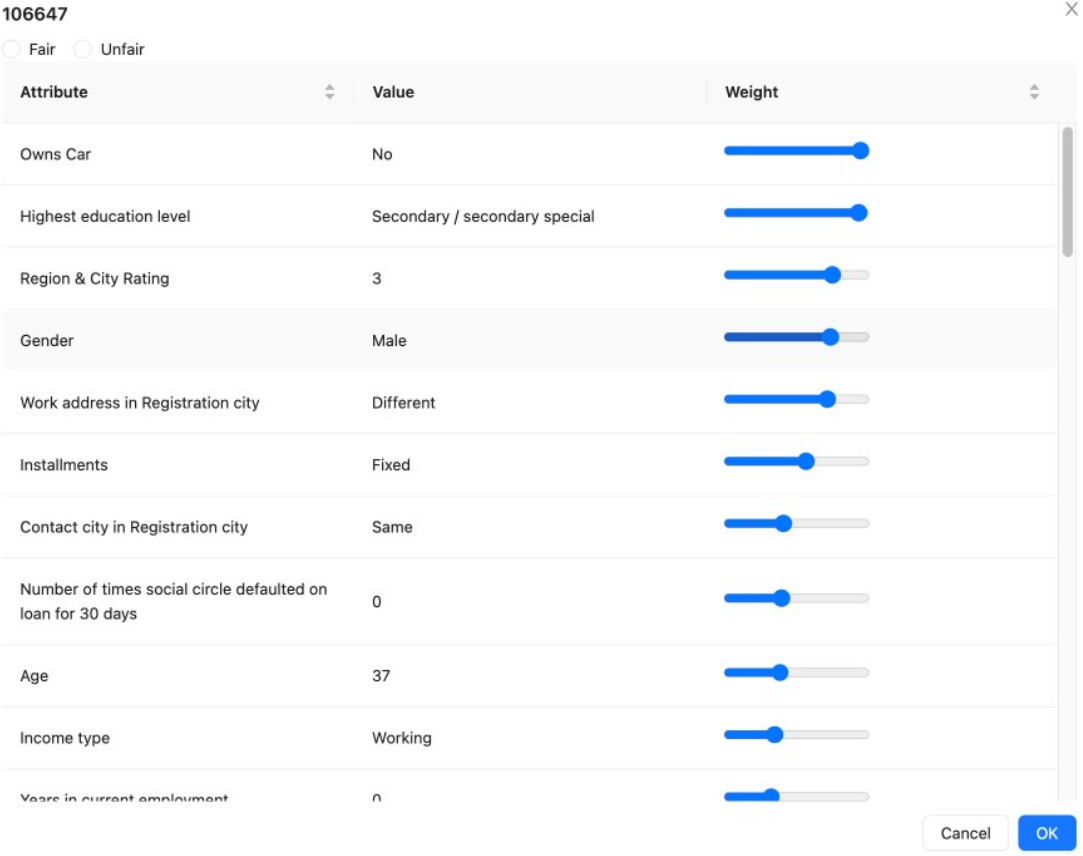}\label{fig:fig1c}}\\
\subfloat[]{%
    \includegraphics[width=0.25\textwidth]{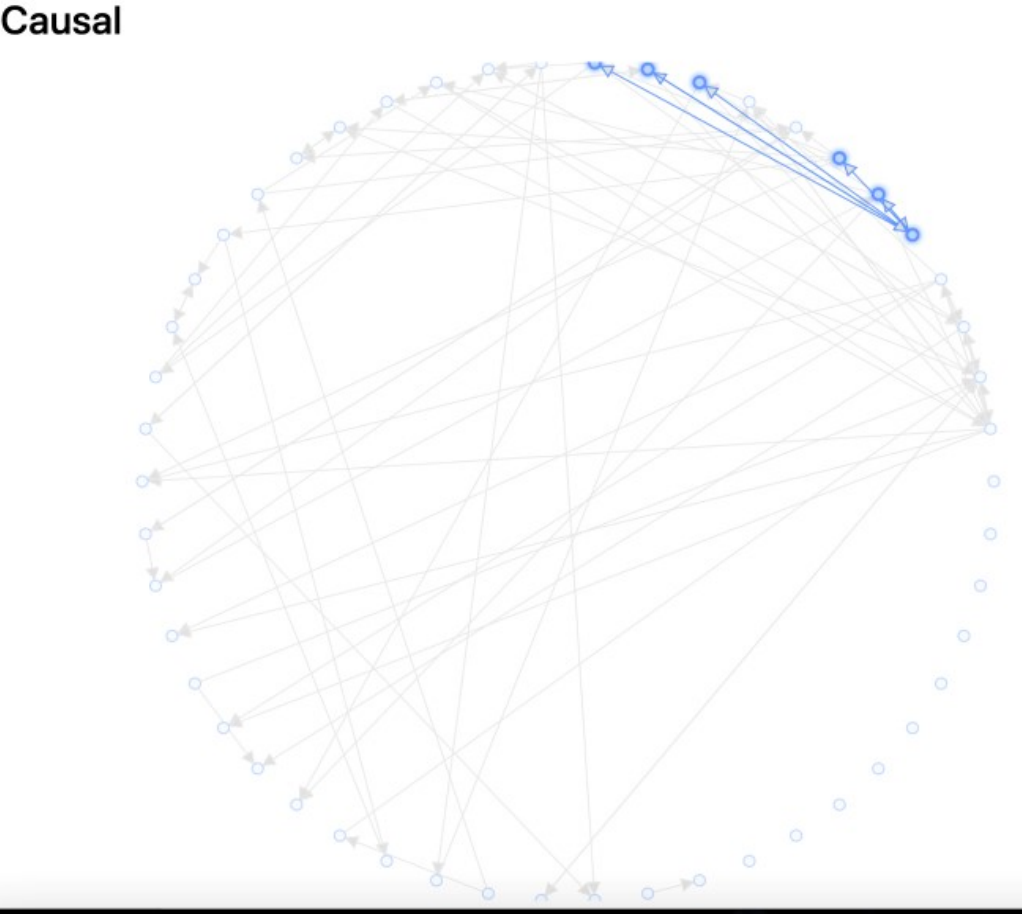}\label{fig:fig1d}}\quad
\subfloat[]{%
    \includegraphics[width=0.25\textwidth]{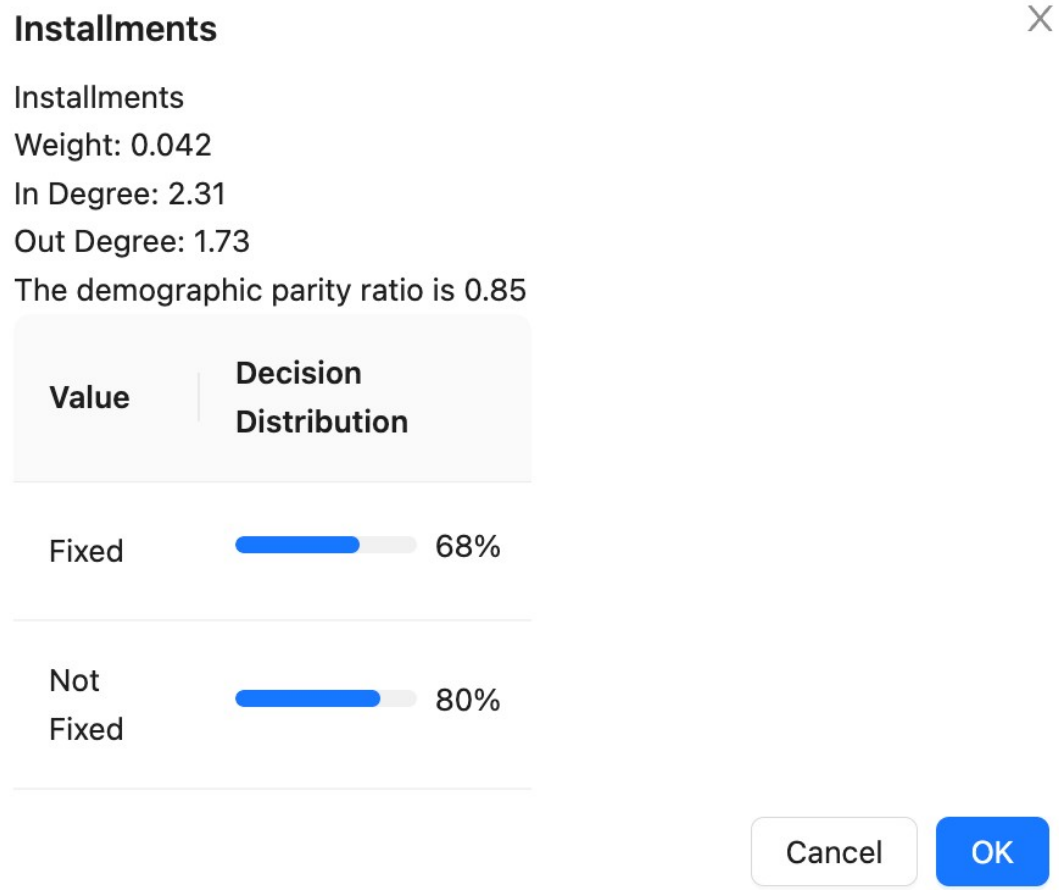}\label{fig:fig1e}}\quad
\subfloat[]{%
    \includegraphics[width=0.18\textwidth]{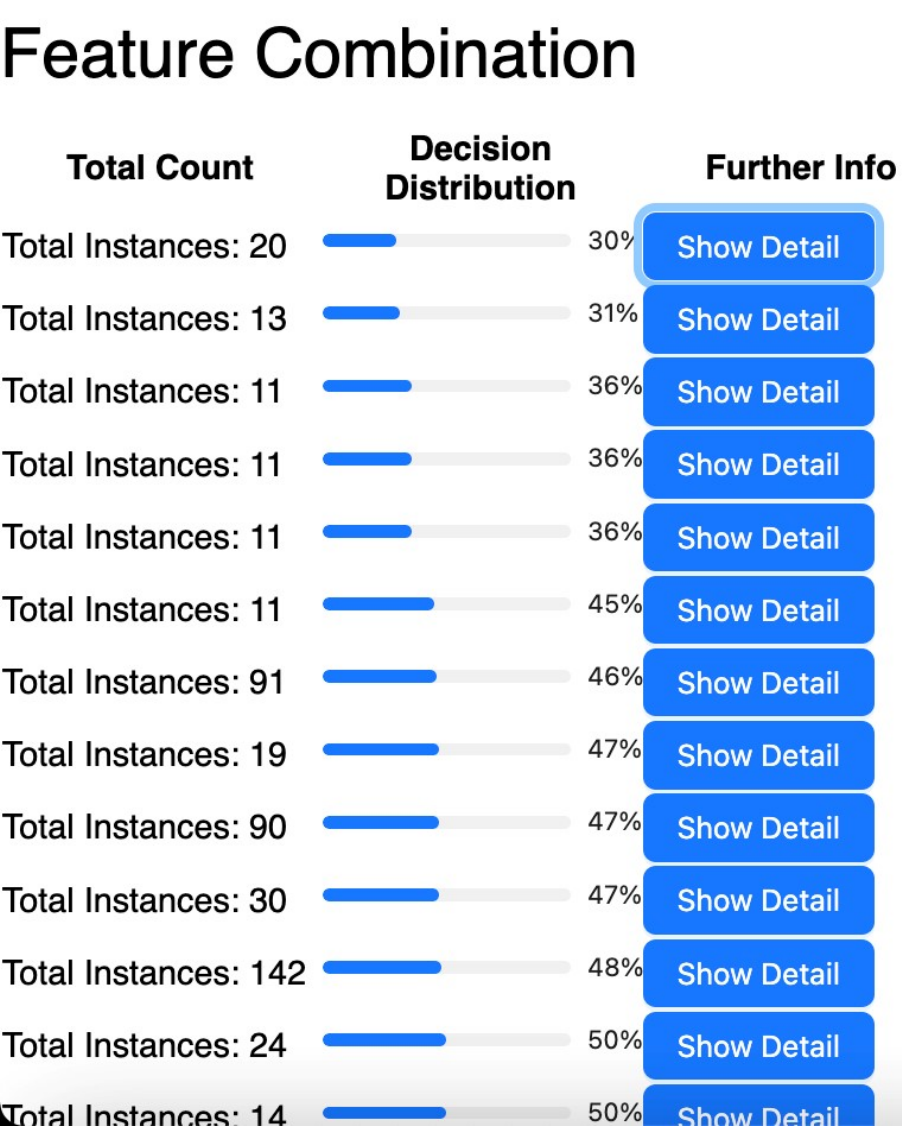}\label{fig:fig1f}}\quad
\subfloat[]{%
    \includegraphics[width=0.25\textwidth]{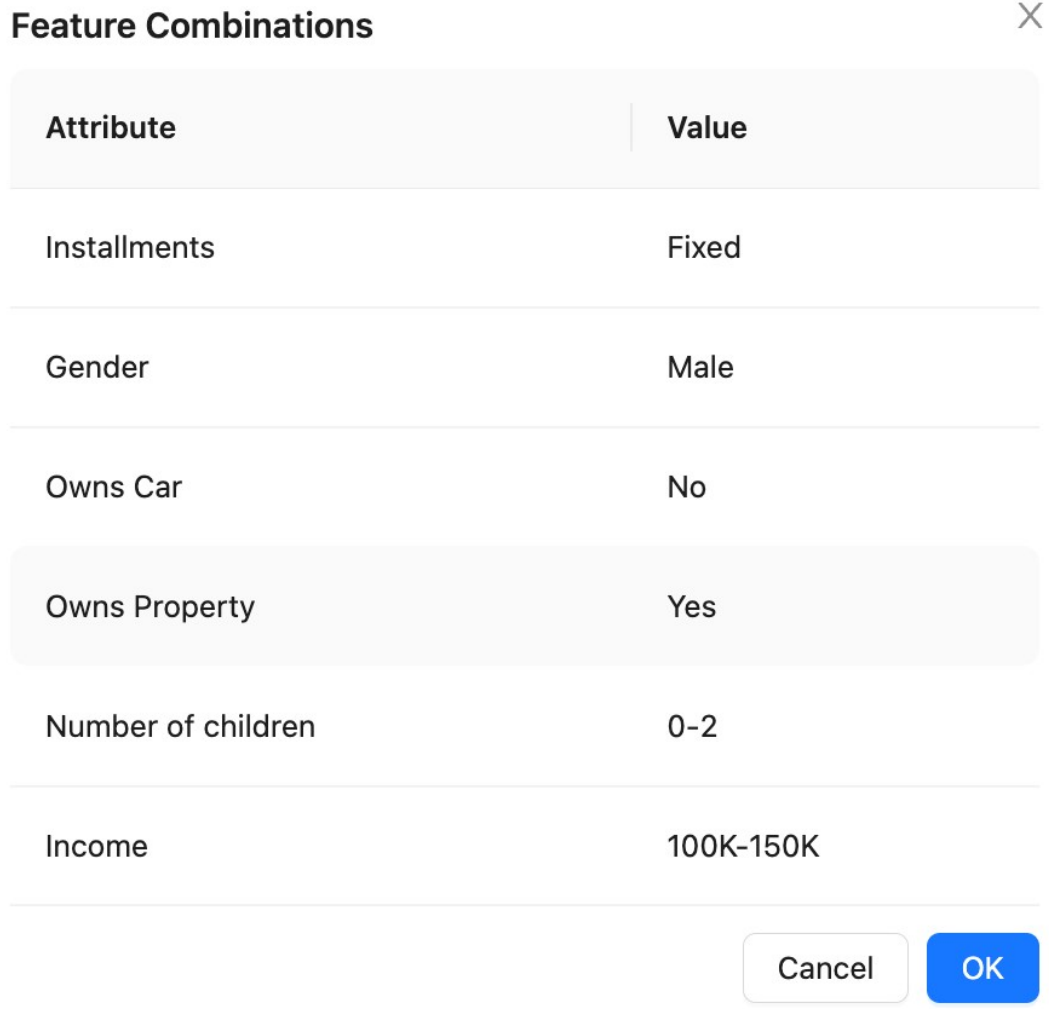}\label{fig:fig1g}}%
\caption{ The UI used in user study 1 to collect participants' feedback on the fairness of the AI model's decisions on the outcome of loan applications. (a) \emph{System Overview}. \emph{Dataset View}: (b) application details list and similarity graph, (c) user feedback input modal showing when the ``Decide'' button of an application is clicked. \emph{Model View}: (d) causal graph of the model's attributes and (e) details of the \texttt{Installment} node shown when the node is clicked. (f) Sorted list of acceptance rates for loan application groups shown below the causal graph and (g) the feature combinations of the group of applications having the lowest acceptance rate.}
\Description{The system overview at the top includes a bar showing acceptance rate, two pie charts showing percentage of accepted or rejected applications marked fair by the user, a button which when pressed gives an explanation of the model, and the Consistency metric. The application list in the second row on the left shows applications by id, predicted decision, and decision confidence. To the right is a similarity graph, a scatterplot of applications along level of similarity to the currently selected application in the application list, and decision confidence. The user can provide feedback through a modal window, which shows attribute names, their value and the weight as an adjustable slider bar. The causal graph shows all attributes as nodes arranged in a circular layout, with causal relationships as lines between them. When the user clicks a node it opens up a window showing details about that attribute. Below the causal graph is a list of feature combination groups, listing total number of instances in this subgroup, the acceptance rate, as a button to get further details. Once a user clicks on this button, they get the subgroup details in terms of the attribute names and values.}
\label{fig:ui_us1}
\end{figure}

The prototype has two main views in separate tabs: the \textbf{dataset view} (Fig.~\ref{fig:ui_us1}b-c) and the \textbf{model view} (Fig. \ref{fig:ui_us1}d-g). Both views show a \emph{system overview} at the top (Fig.~\ref{fig:fig1a}). The system overview component provides overall model information to the user: it  shows the percentage of accepted applications in the test data, and, respectively, the percentage of accepted or rejected application decisions marked as fair or unfair by the user. It also provides a brief explanation of how the model works in lay user-friendly terms, and shows the \emph{Consistency} score of the model. For Consistency, we computed the metric using the 5 most similar applicants for each applicant. 

\textbf{Dataset view:} This view gives more information to the user about individual applications and allows the user to give feedback. All 1000 instances were shown to all participants but they could choose which and how many instances to provide feedback on. This view contains a component which shows a list of applications (Fig.~\ref{fig:fig1b}) with their predicted decisions, i.e. whether a loan application should be accepted or rejected, and prediction confidence of the trained model. This list can be sorted and filtered to find applications to look at and feedback on. 

%When an application is clicked in the list, it is highlighted in the similarity graph and all other applications are shown. Below the similarity graph, we displayed the details of the currently selected application. If a different application is selected in the similarity graph, then its details are presented alongside the currently selected application.
Next to this list view is a similarity graph component showing the similarity between applications on the y-axis and prediction confidence on the x-axis. This provides more contextual information of an application in comparison with other applications. When a user clicks on an application in the list, it highlights it in the similarity graph, revealing all other applications. Below the graph, details of the application selected in the list are displayed. Selecting a different application in the graph presents its details alongside the currently selected application.

Feedback is provided by users through the list view. A user can click on ``Decide'' next to an application, which opens a modal window (Fig.~\ref{fig:fig1c}). In this window,  users are able to mark applications as Fair or Unfair, and adjust the original feature weights used in the XGBoost model to reflect their own views of their importance. Users were able to change their feedback (fairness tag and/or feature weights) for an application with the updated feature weights being stored and shown to them until further changes were made.
%Users can provide feedback by clicking on an application in the list view. This brought up a modal window (Fig.~\ref{fig:fig1c}), where they could mark the application as either Fair or Unfair, as well as sort on the attribute and weight and adjust the weight of the feature through a slider.

\textbf{Model view:} %This component shows a causal graph~\cite{wright1934method,pearl1995causal} of all attributes. The causal graph's edges show the attributes' influence on the decision. Clicking on an attribute's node shows the attribute's weight (from the XGBoost feature weights), in-degree and out-degree, the demographic parity ratio, and a graph showing the acceptance rate for each attribute value (the proportion of applications with the specific attribute value having been accepted). Below the causal graph, a list of acceptance rates for groups of loan applications (with a minimum of 10 instances in a group), sorted by increasing acceptance rate, is shown. Clicking on any of these rates, the feature combinations that result in that rate are shown.
This view presents global model information to the user to find aspects to feedback on. It shows a causal graph~\cite{wright1934method,pearl1995causal} showing all attributes and their influence on decisions (Fig.~\ref{fig:fig1d}). Clicking an attribute's node reveals the feature weight used in the orginal XGBoost model, in-degree and out-degree, demographic parity ratio (DPR), and a graph illustrating acceptance rates for each attribute value (proportion of applications with that value accepted) (Fig.~\ref{fig:fig1e}). For \emph{DPR}, we determined the unprivileged and privileged groups of a protected attribute based on the minimum and maximum selection rates, respectively according to \citet{weerts2023fairlearn}. Below the causal graph, a sorted list of acceptance rates for loan application groups containing at least 10 instances is displayed (Fig.~\ref{fig:fig1f}). Clicking a rate shows the associated feature combinations (Fig.~\ref{fig:fig1g}).

\subsubsection{Participants.} 60 participants over 18 years old were recruited through Prolific (\url{https://www.prolific.com/}), a participant recruitment platform, but 2 were excluded due to data issues. Overall, 58 participants provided feedback (30 female, 28 male) and were compensated with £5 for a 30-minute session. In order to target lay users, participants were asked not to respond if they had a background in AI. The study was approved by the University of Glasgow College of Science and Engineering Ethics Committee.% We make participant data available at \emph{anonymised github}. 

\subsubsection{Procedure.} %On accepting the HIT in Prolific, participants were directed to the study URL. First, the information sheet and consent form were shown. If they accepted, they were directed to the \emph{tutorial}. The tutorial gave an overview of the software and showed them how to use it.
After accepting the advertised task on Prolific, the participants were directed to the study website. After gaining informed consent, participants were exposed to a tutorial for the UI, which they could study for as long as they liked. They then moved on to the main task to interact with the UI. We kept the interaction with the UI to 20 minutes to minimise participant fatigue, after which the study ended. The median time participants spent on the study was 26 minutes and 56 seconds. 

As a main task, they were asked to assess the application decisions, mark up the ones that they considered fair or unfair and adjust feature weights that would, in their view, make the decision fairer. While they were not able to change the label of the prediction, they could change the Fair/Unfair label for a decision as often as they liked, in case they changed their minds. They could not see the effects of their feedback on a re-trained model. 

\subsubsection{Data Collected.} %The collected data consists of a set of log files (one per participant) and a set of demographic data about the participants. All data is anonymous using a unique random Prolific ID to differentiate participants. The log files recorded the applications, feature combinations, causal graph nodes, similar applications participants clicked on, the applications' rating (``fair'', ``unfair'') and weights they proposed, the attributes they filtered, and the applications' prediction confidence sorting (ascend. or descend.). Participants' demographics consist of age, gender, ethnicity, country of birth, country of residence, nationality, language, student status, and employment status. 
The collected data includes participant-specific log files and demographic information passed on from Prolific. Log files captured time-stamped interactions with applications, feature combinations, causal graph nodes, similar applications, fairness ratings, proposed weights, filtered attributes, and sorting on prediction confidence. Participant demographics consist of age, gender, ethnicity, birth and residence countries, nationality, language, student and employment status.

\subsubsection{Offline Experiments}
The collected data and code for the feedback integration and analyses can be found at
\url{https://github.com/evdoxiataka/effi_user_study1}.

Feedback was given through the
UI, as we showed in Section 3.1.2. The feedback was whether an application
is judged fair or unfair, and adjusted feature weights. Our experiments were based on $58$ participants who provided feedback labels (``fair'' or ``unfair''), and $49$ who also provided feedback weights for the trained XGBoost model alongside the feedback label. 

There are potentially many approaches for integrating user feedback into the model. For example, possibilities include: oversampling the feedback instances, feature pruning, or integrating user feedback into a different type of model that uses feature weights directly (e.g. logistic regression). We chose to focus on three approaches because they are relatively straightforward to implement for integrating participants' feedback into XGBoost:
\begin{itemize}
    \item \texttt{Labels}: Only the feedback labels were used to integrate the feedback into the training set and weights were ignored. If the feedback label of a feedback instance was ``unfair'', we flipped its target value and then integrated this feedback instance into the re-training set. If the feedback label was ``fair'', we simply included the feedback instance in the re-training set. 
    \item \texttt{Labels\_Unfair}:  Only the feedback instances labeled as ``unfair''  were integrated into the re-training set after their target value was flipped. We ignored the feedback instances labeled as ``fair'' as it is potentially easier for users to identify problematic instances, i.e. unfair instead of fair ones. 
    \item \texttt{Labels+Weights}: We integrated the feedback instances containing both the label and weights into the re-training set. The feedback labels were treated as in the \texttt{Labels} approach. For the feedback weights, we set the XGBoost\footnote{Python~API: \url{https://xgboost.readthedocs.io/en/stable/index.html}} \textit{feature\_weights} argument of the \textit{fit} method to the normalized feedback weights. 
\end{itemize}  

In all approaches, we adjusted the weights of all training instances to address class imbalances and equalize the contribution of feedback and pre-existing training instances. For all of these approaches, we also explored two further avenues in retraining the XGBoost with participants' feedback: 
\begin{itemize}
    \item \texttt{Global}: We integrated all feedback instances across every participant into the training set and then retrained the XGBoost. This simulates a crowd-sourcing setting and allows us to investigate the overall effect of participants' feedback on the model's fairness. 
    \item \texttt{Personalized}: We trained an independent XGBoost for each participant, as follows: a participant's feedback instances were integrated into the training set one at a time in increasing timestamps, with all previous feedback instances remaining in the training set. The XGBoost was retrained from scratch after each integration step. %XGBoost's Python implementation~\footnote{} does not support continuous training over one training instance at a time, 
\end{itemize}

We then evaluated the impact of incorporating participants' feedback on model fairness for each of the different integration approaches, using common group and individual fairness metrics. Five group fairness metrics -- \emph{demographic parity ratio (DPR), conditional demographic disparity (CDD), equal opportunity difference (EOD), average odds difference (AOD), predictive parity difference (PPD)} --  were computed on the 3 protected attributes (\emph{Gender, Marital Status, Age}). Two individual fairness metrics  -- \emph{Consistency (Cons.) and Theil index (TI)} -- were also employed.
We also calculated the \emph{accuracy} of each model.

\begin{figure}[!h]
%  \begin{minipage}[b]{.65\linewidth}
    \centering    \includegraphics[width=0.78\linewidth]{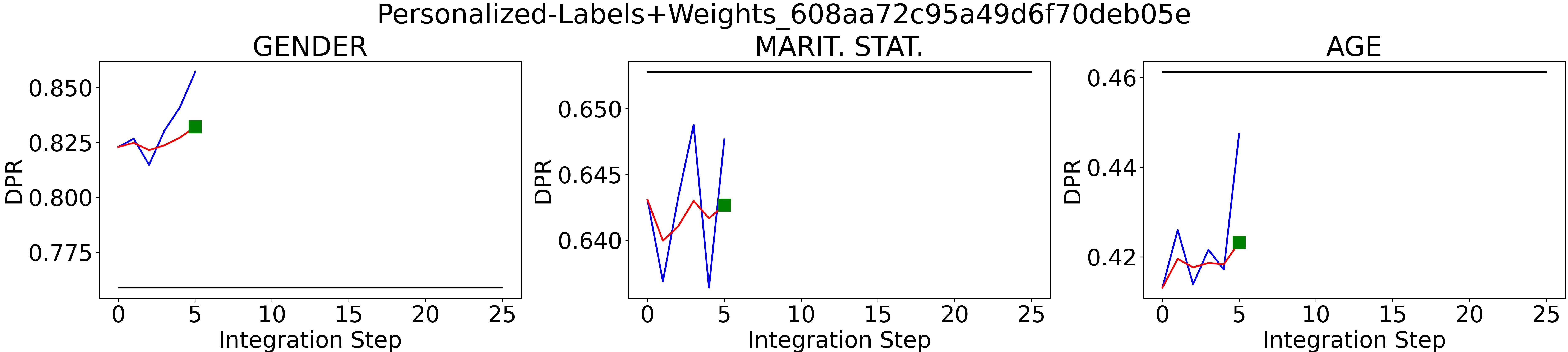}
    \captionof{figure}{DPR of the protected attributes for one of the participants calculated based on the \texttt{Personalized\_Labels+Weights} approach. Black lines show the baseline, blue lines show the raw values, and red lines the CMA values in each integration step. An integration step represents a single feedback instance integration. The green markers represent the CMA points we use to calculate the percentage changes from baseline. It shows that raw feedback (blue) is very noisy which can be smoothed out through the CMA (red) to determine a general pattern of feedback. This participant provided only 6 feedback instances.}
    \Description{The figure shows 3 graphs: 1 each for DPR on gender, marital status and age, calculated based on the \texttt{Personalized\_Labels+Weights} approach, based on the 6 feedback iterations submitted by an individual participant. It shows that raw feedback (blue) is very noisy which can be smoothed out through the CMA (red) to determine a general pattern of feedback. This participant improved DPR for gender against the baseline and deteriorated it for marital status and age.}
    \label{fig:part_graphs}
\end{figure}

To investigate the impact of integrating the feedback in personalized models, metrics were computed on the re-trained model after every integration step. While the global model had a single integration step, the personalized models had multiple integration steps, one per participant feedback instance, resulting in high fluctuations of the metrics as participants provided more feedback (e.g. Fig.~\ref{fig:part_graphs} blue line). Thus, we computed the \emph{cumulative moving average (CMA)} to obtain a feedback trend (e.g. Fig.~\ref{fig:part_graphs} red line); as a participant provides a new feedback instance, the average value of a fairness metric at the time of the feedback instance is calculated for all feedback instances up to that point. The fairness metrics' values based on the CMA in the last integration step were used to calculate the percentage change from the baseline. For example, Fig.~\ref{fig:part_graphs} presents the DPR values for the protected attributes for one of the participants for the \texttt{Personalized\_Labels+Weights} approach. %, all of them labelled as ``unfair''. 

\subsection{Results}

\subsubsection{Baseline}
We computed the baseline for each fairness metric based on the initial trained model shown to participants and used it to compare the results from the re-trained models. Table~\ref{tab:baseline_us1} shows baseline values for fairness metrics we calculated, showing their respective ``ideal'' values. Fairness issues already exist in DPR, EOD, and AOD for protected features, and Consistency before feedback integration. Additionally, there is room for improvement for other fairness measures, such as CDD, PDD, and TI. 

\begin{table*}[!h]
\centering
\caption{Baseline metrics for the XGBoost model used in user study 1 with corresponding ``ideal'' values in parentheses.  The direction of improvement for a metric is shown in parentheses: $\uparrow$ indicates higher better, $\downarrow$ means lower better. All values are rounded to the second significant digit.}
\label{tab:baseline_us1}
    \resizebox{0.6\textwidth}{!}{%
 \begin{tabular}{r|ccccc|}
    \cline{2-6}
    & \multicolumn{5}{c|}{Baseline in User Study 1}\\
    \cline{2-6}
    
    & \textbf{Acc. $\%$}  ($\approxeq 100$)($\uparrow$) & \textbf{Cons.} ($\approxeq 1$) ($\uparrow$) & \textbf{TI} ($\approxeq 0$) ($\downarrow$) &  & \\
    
     & 66.59 & 0.67 & 0.30 & & \\
     
      & \textbf{DPR} ($\approxeq 1$) ($\uparrow$) & \textbf{CDD} ($\le 0$) ($\downarrow$) & \textbf{EOD} ($\approxeq 0$) ($\downarrow$) & \textbf{AOD} ($\approxeq 0$) ($\downarrow$) & \textbf{PPD} ($\approxeq 0$) ($\downarrow$) \\
      
    \textbf{Gender} & 0.76 & -0.04 & 0.14 & 0.15 & 0.02 \\
    \textbf{Marital Status} & 0.65 & -0.04& 0.23 & 0.27 & 0.04 \\
    \textbf{Age} & 0.46 & \num{4e-4} & 0.39 & 0.43 & 0.07\\    
    \cline{2-6}%
\end{tabular}%
}
\end{table*}

\subsubsection{Effects of Integrating Feedback on \texttt{Global} Model Fairness}
We expected this analysis to help us choose fairness metrics that best reflected feedback across all users: either a metric would consistently improve across all attributes (i.e., that metric gives a consistently better signal for fairness for all attributes), or that there would be improvements across all metrics for certain attributes (i.e., that there is an overwhelming signal for an attribute to be considered for fairness). Table~\ref{tab:Global_results_us1_all} provides an overview of the effects of integrating participants' feedback into the accuracy and fairness metrics in the global model. 

\begin{table*}[!h]
\centering
\caption{Metrics for the retrained \texttt{Global} Xgboost model after integrating feedback instances provided for any application. Percentage changes $\%$ from baseline are given in parentheses. Improvements against the baseline are highlighted with a shaded background, light gray =<5\%, dark grey >5\%.}
\label{tab:Global_results_us1_all}
    \resizebox{\textwidth}{!}{%
 \begin{tabular}{r|ccccc|ccccc|ccccc|}

    \cline{2-16}
    
     & \multicolumn{5}{c|}{\texttt{Global\_Labels}} & \multicolumn{5}{c|}{\texttt{Global\_Labels\_Unfair}} & \multicolumn{5}{c|}{\texttt{Global\_Labels+Weights}}\\
     
     \cline{2-16}
      & \textbf{Acc. $\%$} ($\uparrow$) & \textbf{Cons.} ($\uparrow$) & \textbf{TI} ($\downarrow$) & & & \textbf{Acc. $\%$} ($\uparrow$) & \textbf{Cons.} ($\uparrow$) & \textbf{TI} ($\downarrow$) & &  & \textbf{Acc. $\%$} ($\uparrow$) & \textbf{Cons.} ($\uparrow$) & \textbf{TI} ($\downarrow$) &  & \\
      
     & \makecell{66.11\\(-0.73)} & \cellcolor{gray!25}\makecell{0.68\\(1.09)} & \makecell{0.30\\(1.07)} & & & \makecell{64.95\\(-2.47)} & \makecell{0.67\\(-1.17)} & \makecell{0.32\\(6.87)} & & & \makecell{65.53\\(-1.60)} & \cellcolor{gray!25}\makecell{0.68\\(0.53)} & \makecell{0.32\\(5.38)} & & \\
     
      & \textbf{DPR} ($\uparrow$) & \textbf{CDD} ($\downarrow$) & \textbf{EOD} ($\downarrow$) & \textbf{AOD} ($\downarrow$) & \textbf{PPD} ($\downarrow$) & \textbf{DPR} ($\uparrow$) & \textbf{CDD} ($\downarrow$) & \textbf{EOD} ($\downarrow$) & \textbf{AOD} ($\downarrow$) & \textbf{PPD} ($\downarrow$) & \textbf{DPR} ($\uparrow$) & \textbf{CDD} ($\downarrow$) & \textbf{EOD} ($\downarrow$) & \textbf{AOD} ($\downarrow$) & \textbf{PPD} ($\downarrow$)\\
      
    \textbf{Gender} & \cellcolor{gray!25}\makecell{0.77\\(1.71)} & \makecell{-0.04\\(5.60)} & \cellcolor{gray!50}\makecell{0.13\\(-5.77)} & \cellcolor{gray!50} \makecell{0.14\\ (-5.80)} & \makecell{0.02\\(23.41)} & \cellcolor{gray!25}\makecell{0.77\\(1.56)} & \makecell{-0.04\\(8.66)} & \cellcolor{gray!25}\makecell{0.14\\(-4.31)} & \cellcolor{gray!50}\makecell{0.14\\(-9.72)} & \makecell{0.03\\(58.87)} & \cellcolor{gray!25}\makecell{0.77\\(1.71)} & \makecell{-0.04\\(9.24)} & \cellcolor{gray!50}\makecell{0.13\\(-5.81)} & \cellcolor{gray!50}\makecell{0.14\\(-10.78)} & \makecell{0.02\\(44.60)}\\
    
    \textbf{Marit. Stat.} & \cellcolor{gray!25}\makecell{0.66\\(1.77)} & \cellcolor{gray!25}\makecell{-0.04\\(-1.26)} & \cellcolor{gray!50}\makecell{0.21\\(-9.30)} & \makecell{0.28\\(3.24)} & \cellcolor{gray!25}\makecell{0.04\\(-3.99)} & \cellcolor{gray!25}\makecell{0.68\\(4.26)} & \makecell{-0.04\\(1.70)} & \cellcolor{gray!50}\makecell{0.19\\(-15.05)} & \cellcolor{gray!50}\makecell{0.24\\(-9.42)} & \makecell{0.04\\(10.78)} & \cellcolor{gray!25}\makecell{0.67\\(1.97)} & \cellcolor{gray!25}\makecell{-0.04\\(-2.40)} & \cellcolor{gray!50}\makecell{0.19\\(-14.71)} & \cellcolor{gray!25}\makecell{0.26\\(-3.67)} & \cellcolor{gray!50}\makecell{0.04\\(-8.41)}\\
    
    \textbf{Age} & \cellcolor{gray!25}\makecell{0.47\\(1.88)} & \cellcolor{gray!50} \makecell{\num{-4e-4}\\(-199.38)} & \cellcolor{gray!25}\makecell{0.38\\(-2.03)} & \cellcolor{gray!25}\makecell{0.43\\(-1.68)} & \makecell{0.08\\(10.33)} & \cellcolor{gray!25}\makecell{0.47\\(2.31)} & \makecell{\num{1e-3}\\(160.98)} & \cellcolor{gray!25}\makecell{0.38\\(-3.38)} & \cellcolor{gray!25}\makecell{0.43\\(-1.81)} & \makecell{0.08\\(23.09)} & \cellcolor{gray!25}\makecell{0.47\\(1.03)} & \makecell{\num{2e-3}\\(404.31)} & \cellcolor{gray!25}\makecell{0.38\\(-2.77)} & \makecell{0.44\\(0.14)} & \makecell{0.07\\(2.17)}\\
    \cline{2-16}%
\end{tabular}%
}
\end{table*}

Overall, we note that all integration approaches drop in accuracy, however, this decrease in Accuracy is not very large. The largest decrease in Accuracy (-2.47\%) is with the \texttt{Global\_Labels\_Unfair} approach taking all feedback instances into account, dropping from 66.59 to 64.95. Previous work \cite{mehrabi2022} has noted that bias mitigation often results in Accuracy decreases, and that accuracy and fairness need to be appropriately balanced.

We also found that individual fairness metrics mostly deteriorated compared to the baseline across all approaches, except for Consistency in the \texttt{Global\_Labels} approaches and \texttt{Global\_Labels+Weights} approaches. Again, these improvements compared to the baseline are relatively small, on average 0.59\%.

We then turned to group fairness. Our analysis showed that when integrating all feedback instances, there are some fairness metrics that consistently show improvements across \emph{all} protected features and \emph{all} integration approaches (highlighted cells in Table~\ref{tab:Global_results_us1_all}). DPR improves 2.02\% while EOD improves 7.01\%, on average. AOD improves for \emph{all} attributes only in the \texttt{Global\_Labels\_Unfair} approach, but also increases fairness on Age in the \texttt{Global\_Labels} approach and Marital Status in the \texttt{Global\_Labels+Weights}, with 10.67\% improvements on average. We found that the improvements for DPR, EOD and AOD seem to be larger in the \texttt{Global\_Labels\_Unfair} approach.

\subsubsection{Effects of Integrating Feedback on \texttt{Personalized} Model Fairness}
\begin{figure}[!h]
\begin{minipage}[b]{\linewidth}
\centering
 \captionof{table}{Average values of DPR, EOD, and AOD for the personalized models of all participants in user study 1.}
\label{tab:mean_personalized}
    \resizebox{0.66\textwidth}{!}{%
     \begin{tabular}{r|ccc|ccc|ccc|}
    
    \cline{2-10}
    
     & \multicolumn{3}{c|}{\texttt{Personalized\_Labels}} & \multicolumn{3}{c|}{\texttt{Personalized\_Labels\_Unfair}} & \multicolumn{3}{c|}{\texttt{Personalized\_Labels+Weights}}\\
     
     \cline{2-10}
     
      & \textbf{DPR} ($\uparrow$) & \textbf{EOD} ($\downarrow$) & \textbf{AOD} ($\downarrow$) & \textbf{DPR} ($\uparrow$) & \textbf{EOD} ($\downarrow$) & \textbf{AOD} ($\downarrow$) & \textbf{DPR} ($\uparrow$) & \textbf{EOD} ($\downarrow$) & \textbf{AOD} ($\downarrow$) \\
      
    \textbf{Gender} & \cellcolor{gray!25}\makecell{0.76\\(0.37)} & \cellcolor{gray!25}\makecell{0.14\\(-1.23)} & \cellcolor{gray!25}\makecell{0.15\\(-3.35)} & \cellcolor{gray!25}\makecell{0.76\\(0.27)} & \cellcolor{gray!25}\makecell{0.14\\(-0.86)}& \cellcolor{gray!25}\makecell{0.15\\(-2.85)}& \cellcolor{gray!25}\makecell{0.77\\(1.75)}& \cellcolor{gray!50}\makecell{0.13\\(-6.39)}& \cellcolor{gray!50}\makecell{0.14\\(-8.44)}\\
    
    \textbf{Marit. Stat.} & \makecell{0.65\\(-0.63)} & \cellcolor{gray!25}\makecell{0.22\\(-2.07)} & \makecell{0.28\\(2.92)}& \makecell{0.65\\(-0.36)}& \cellcolor{gray!25}\makecell{0.22\\(-3.44)}& \makecell{0.28\\(2.21)}& \makecell{0.65\\(-0.84)}& \cellcolor{gray!25}\makecell{0.22\\(-2.13)}& \makecell{0.28\\(2.94)}\\
    
    \textbf{Age} & \makecell{0.46\\(-1.24)} &  \makecell{0.40\\(1.52)} & \makecell{0.44\\(1.48)}& \makecell{0.45\\(-1.69)}& \makecell{0.40\\(1.90)}& \makecell{0.44\\(1.76)}& \makecell{0.45\\(-2.71)}& \makecell{0.40\\(3.16)}& \makecell{0.45\\(2.93)}\\
    \cline{2-10}%
\end{tabular}%
}
\end{minipage}\\
\begin{minipage}[b]{\linewidth}
\centering
\includegraphics[width=0.6\linewidth]{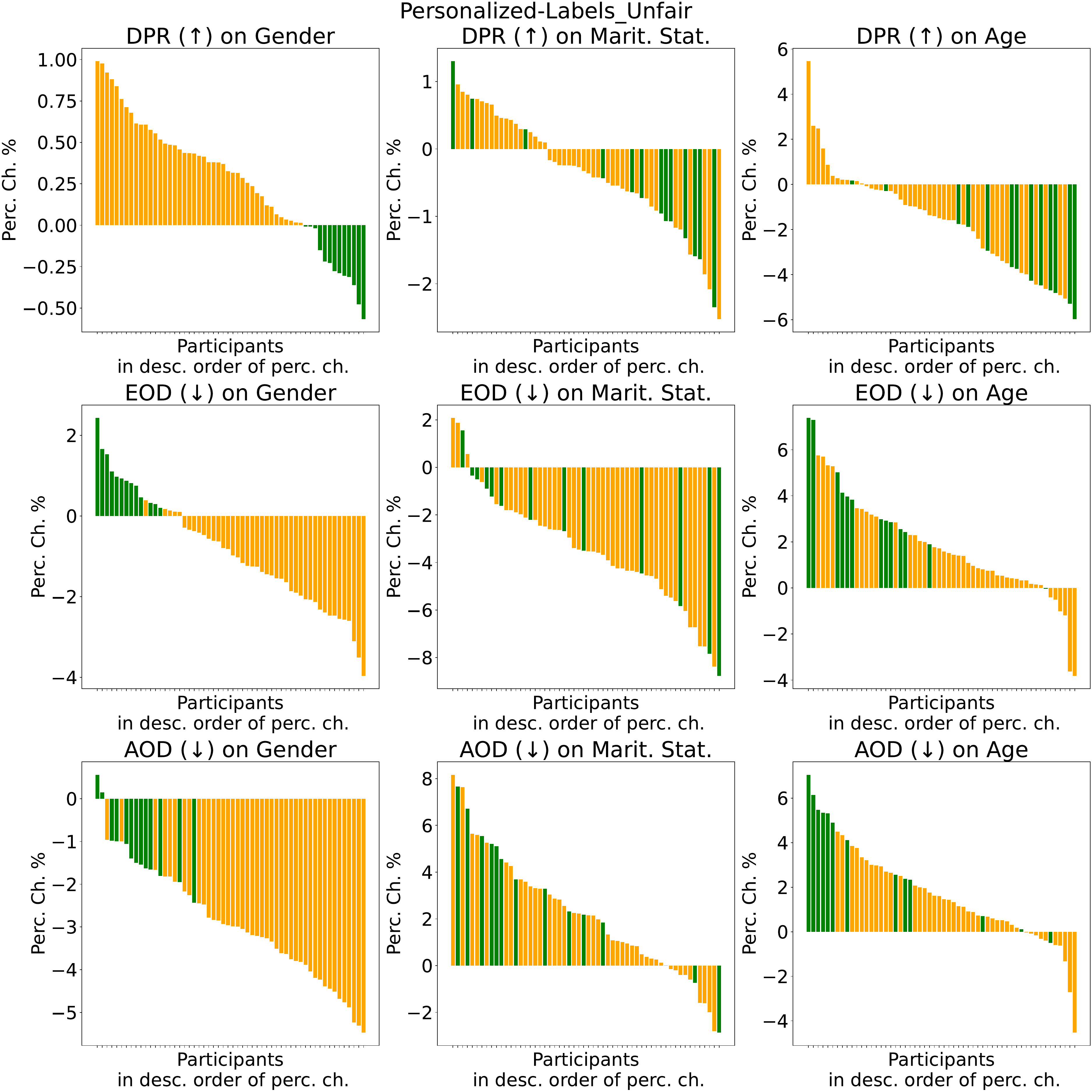}%
\captionof{figure}{Percentage change of DPR, EOD, AOD from baseline per participant for the \texttt{Personalized\_Labels\_Unfair} approach models presented in bar graphs. Participants who deteriorated DPR on Gender are show in green. For DPR, over half the participants improved the metric for Gender, but many fewer improved it for Marital Status and Age. For EOD, over half the participants improved the metric on Marital Status, but very few did on Gender and Age. For AOD, almost all participants improved the metric on Gender (more than for DPR), but very few improved the metric for Marital Status and Age.}
\Description{The figure shows 9 bar graphs for the \texttt{Personalized-Labels\_Unfair} feedback integration method; 3 for DPR, 1 each for gender, marital status and age, 3 for EOD, again, 1 each for gender, marital status and age, and 3 for AOD, again 1 each for gender, marital status and age. It shows that for DPR over half the participants improved the metric for gender, but many fewer improved it for marital status and age. For EOD, over half the participants improved the metric on marital status, but very few did on gender and age. For AOD, almost all participants improved the metric for gender (more than for DPR), but very few improved the metric for marital status and age.}
\label{fig:personalized}
\end{minipage}
\end{figure}

We wanted to investigate how the fairness evolved based on the feedback of individual participants in their personalized models. We focused on DPR, EOD, and AOD as these already had shown good results in the global model. Table~\ref{tab:mean_personalized} shows the DPR, EOD and AOD values and changes for each protected attribute across all participants. The results are mainly in line with what we found in the global model, however, the improvements on average are smaller: we found that fairness on Gender was improved on average by 0.98\%  in \texttt{Personalized\_Labels}, and it improved by 2.15\% on average in the \texttt{Personalized\_Labels+Weights} approach. For Age, there was also a consistent pattern of slight improvements across all participants in the \texttt{Personalized\_Labels} and the \texttt{Personalized\_Labels\_Unfair} approaches. For Marital Status, there was only a consistent improvement in the \texttt{Personalized\_Labels\_Unfair} approach. 

Fig.~\ref{fig:personalized} presents the percentage change of DPR, EOD, and AOD from baseline computed for each participant's model, sorted by the amount of percentage change. Focusing on DPR for Gender, we found that about a quarter of participants gave feedback which deteriorated fairness (shown in green). Most of these individuals also deteriorated DPR on the other protected values, along with about quite a substantial proportion of other participants. For EOD, a large proportion of participants improved fairness on Gender and also on Marital Status, but only few improved AOD on Age. For AOD, almost all participants managed to achieve substantial improvements to the fairness metric on Gender but also most participants deteriorated AOD on Marital Status and Age. Note that participants who deteriorated DPR on Gender appear to often deteriorate other metrics and attributes as well. 

It could be that the feedback participants gave and the improvements that they made was influenced by their demographic background, for example, the participants' gender might influence whether they focused on Gender in fairness. We therefore ran a general linear model on participants’ age, gender, and ethnicity with the protected attributes' DPR and AOD for their personalized model but none of these demographic variables nor their interactions affected these fairness metrics. 

\subsubsection{Other Feedback}
We have already shown that the feedback participants gave for fairness metrics on protected attributes is very complex. Another reason for these complex patterns might be that they also focused on non-protected attributes, and used very different fairness notions to what we captured e.g. procedural fairness and ``correctness''. To explore this, we turned to the interaction logs to investigate how much participants interacted with each attribute and what changes they suggested to the weights of the attributes. Table~\ref{tab:attributes_interaction_us1} shows the top 10 attributes that participants interacted with, along with the average proposed weight change. (Details for all attributes can be found in the supplementary material.)

We found that Age and Gender featured quite highly in terms of filtering applications and that Gender, Marital Status, and Age were also in the top 10 attributes where participants suggested a weight change. However, less than half of participants interacted with any of them. Instead, other, non-protected attributes which are usually not considered in fairness metrics were frequently adjusted. For example, the attribute for which most participants proposed a weight change was \texttt{Owns Car} (applicant's ownership of a car). We also noted that a number of the attributes that participants interacted with deal with employment, income, owning a home, etc. Similar to previous work~\cite{nakao2022}, participants might have considered \emph{affordability} as a fairness notion.

%It is interesting to investigate how consistently participants adopted this position all through their feedback instances and to one another. 
\subsubsection{Interactions with the UI Components}
We found that only about a quarter of the participants clicked on nodes in the causal graph and only about a third of them selected to view similar applications in the scatter plot. This could suggest that participants did not find these UI components very useful. Indeed, lack of use could explain our results in terms of individual fairness. However, more than half of the participants clicked to view the feature combinations below the causal graph, possibly indicating an interest in (sub)group fairness.

\begin{table}[!t]
    \centering   
    \caption{Top 10 attributes that participants interacted with in at least one of their feedback instances, along with the average proposed weight change in comparison to model's weights before integrating the feedback. It shows that protected attributes such as Gender, Marital Status and Age are in the top 10 but that there are other attributes which also seemed to be important to participants.}
    \label{tab:attributes_interaction_us1}
    \rowcolors{2}{gray!25}{white}
    \resizebox{0.68\linewidth}{!}{%
 \begin{tabular}{rc|rcc}
    \hline
    \multicolumn{2}{c|}{\textbf{Filtering for Attribute}} & %\multicolumn{2}{c|}{\textbf{Clicks on Attribute Nodes}} & 
    \multicolumn{3}{c}{\textbf{Attribute Weight Changes}}\\
     \hline
      \textbf{Attribute} & \makecell[r]{\textbf{No.} \textbf{Part.}} & 
      %\textbf{Attribute} & \makecell[r]{\textbf{No.}\\ \textbf{Part.}} & 
      \textbf{Attribute} & \makecell[r]{\textbf{No.} \textbf{Part.}} & \makecell[r]{\textbf{Av. Weight} \textbf{Ch. $\%$}}\\
   \hline
    \textbf{Age} & 26 & 
    %\makecell[r]{Number of \\family members} & 5 & 
    \text{Owns Car} & 39 & -62.59\\

    Income & 14 & 
    % Installments & 4 & 
    \textbf{Gender} & 28 & -72.95\\

   \textbf{Gender} & 14 & 
   % Application Hour & 3 & 
   \text{Income type} & 27 & 113.11 \\

    Owns Car & 10 &  
    % Income & 3 & 
    \makecell[r]{Years in  current employment} & 25 & 104.94\\

    \makecell[r]{Accompanied while applying} & 9 &
    % Loan Credit amount & 2 & 
    \text{Highest education level} & 24 & -51.70\\

    Occupation Type & 6 & 
    % \makecell[r]{Years since \\changing registration} & 2 & 
    \text{Housing situation} & 24 & 165.83\\

    Application Day & 6 & 
    % Application Day & 2 & 
    \text{Owns Property} & 23 & 283.42\\

    Owns Property & 5 & 
    % Goods Price & 2 & 
    \textbf{Age} & 23 & 62.75\\

    \makecell[r]{Contact address located \\ in Registration region} & 5 & 
    % Has Employee Phone & 2 & 
    \text{Income} & 22 & 312.58\\

    Highest education level & 5 & 
    % \makecell[r]{Contact address located \\ in Registration region} & 2 & 
    \textbf{Marital Status}  &  22 & 133.74\\
    \hline
\end{tabular}%
}
\end{table}

\subsection{Implications}
These results suggest four implications. First, AOD was the metric which led to the most improvements on Gender for the most participants, across both global and personalized models. We argue that this would be a good metric to choose to obtain feedback on and evaluate fairness on this attribute. Second, while most participants seem to have tried to improve the fairness of Gender, both on DPR and AOD, they did not improve Marital Status or Age for these metrics. This could mean that either 1) they did not care about these other attributes or 2) that they might have used a different fairness metric, e.g. EOD on Marital Status. Further research is necessary to understand users' intentions behind their feedback. Third, it appears that some participants do not provide feedback that improves fairness for \emph{any} metric on \emph{any} protected attribute. How to identify these individuals and how to treat their feedback is an open research area. Last, feedback might have had unintended effects on the fairness of attributes. Recall that we did not show them how their feedback affects the fairness metrics because we collected user feedback \emph{outside} of a true IML setting: we did not retrain a participant's personalized model online, nor did we show the updated predictions of the retrained model to participants. Hence, it could be argued that participants were ``flying blind'' as they were not given any information about the effect of their feedback on the model's fairness. Thus, we conducted a second user study to remedy this.

\section{User Study 2}
User study 2 followed up and extended user study 1. Similar to the previous study, we elicited feedback on the fairness of loan application decisions produced by an AI model but this time we integrated participants' feedback into a personalized model online and showed them the impact of the retrained model. We then also constructed a global model in an offline experiment.

\subsection{Method}
Because this study is very similar to study 1, we only highlight the major differences in our methods. 

\subsubsection{The Dataset and AI model}%, Dataset, and AI Model}
The code for preparing the data and training the model initially can be found in \url{https://github.com/evdoxiataka/effi_user_study2}. 

%A commonly employed dataset is the German Credit dataset \cite{germanCredit1994}, however, it has been noted that this dataset is relatively small, with few features, and does not mirror real data.    \citet{nakao2022} did not use an open-source dataset to enable replication of the findings. We followed \citet{taka2023} in using the Home Credit open-source loan application dataset \cite{homeCreditDataset}.  
We again used the Home Credit open-source loan application dataset~\cite{homeCreditDataset} with the same $49$ features.  %Similarly to \citet{taka2023}, we inversed the binary target variable of the dataset noting clients with payment difficulty to predict the acceptance or rejection of a client's application. 
In this study we only used the labeled training set, S1: we sampled $11$K instances while preserving the original distribution of the target variables\footnote{We used the \texttt{StratifiedShuffleSplit} function of the Python  \texttt{sklearn} library.} %using the for training and testing to reduce the computation of model training, 
and another $100$ to show to participants during the study. We used $10$K instances as training set and the remaining $1$K instances for a test set. During the user study, we presented $100$ loan applications in the test set to participants through the UI. As before, we performed pre-processing on the selected features, including missing values imputation, normalization of specific features, and hot encoding of categorical variables.  %We used the same $49$ features with \citet{taka2023} for training. 
%For instance selection, we sub-sampled the $100$K instances \citet{taka2023} used for the training and validation of the model down to $11,1$K by preserving the distribution of the Target variable. From this subset, we created a $10$K training set to train the AI model initially, a $1$K validation set to evaluate the accuracy and fairness of the model before and after integrating participants' feedback, and a $100$-instances test set shown to participants through the UI. 

%Reducing the size of the training and test sets was necessary to achieve retraining the model and provide users with information about the impact of their feedback in real time. Any other data pre-processing we conducted (i.e., missing values imputation, normalization of certain features, hot encoding of categorical variables) is in line with \citet{taka2023}.
As in study 1, we used a XGBoost decision tree classifier~\cite{XGBoost, Chen2016}. The model was trained \emph{offline} on the training set addressing class imbalances. %Home Credit dataset training subset of $10$K instances that we formed as described above. 
As participants were interacting and providing feedback, retraining of the model took place \emph{online} and in real-time as a new feedback instance by a participant was recorded. Participants were only able to label applications as unfair and/or suggest weight changes. The unfair labeled applications were integrated into the training set incrementally by flipping its target value like we did in the offline experiments in study 1 and the model was retrained on the extended training set, resulting in a different model for each participant, using the \texttt{Labels\_Unfair} approach. Unlike study 1, if only weight changes were provided, the model was retrained by setting the attribute weights to participant's weight feedback but without integrating the application label into the extended training set; we call this the \texttt{Labels\_Unfair+Weights} approach. 

Information provided to participants as the model was updated included the classification accuracy, Demographic Parity Ratio (DPR) and Average Odds Difference (AOD), using the minimum and maximum groups if there were more than two groups (see section 4.1.2 Fairness Metrics). % because participants in \citet{taka2023} seemed to have improved these the most and we wanted to avoid overwhelming participants with many different notions of fairness.
%Class weights were used in the training to deal with class imbalance. 

\subsubsection{The User Interface} 

\begin{figure}
\subfloat[]{%
    \includegraphics[width=\textwidth]{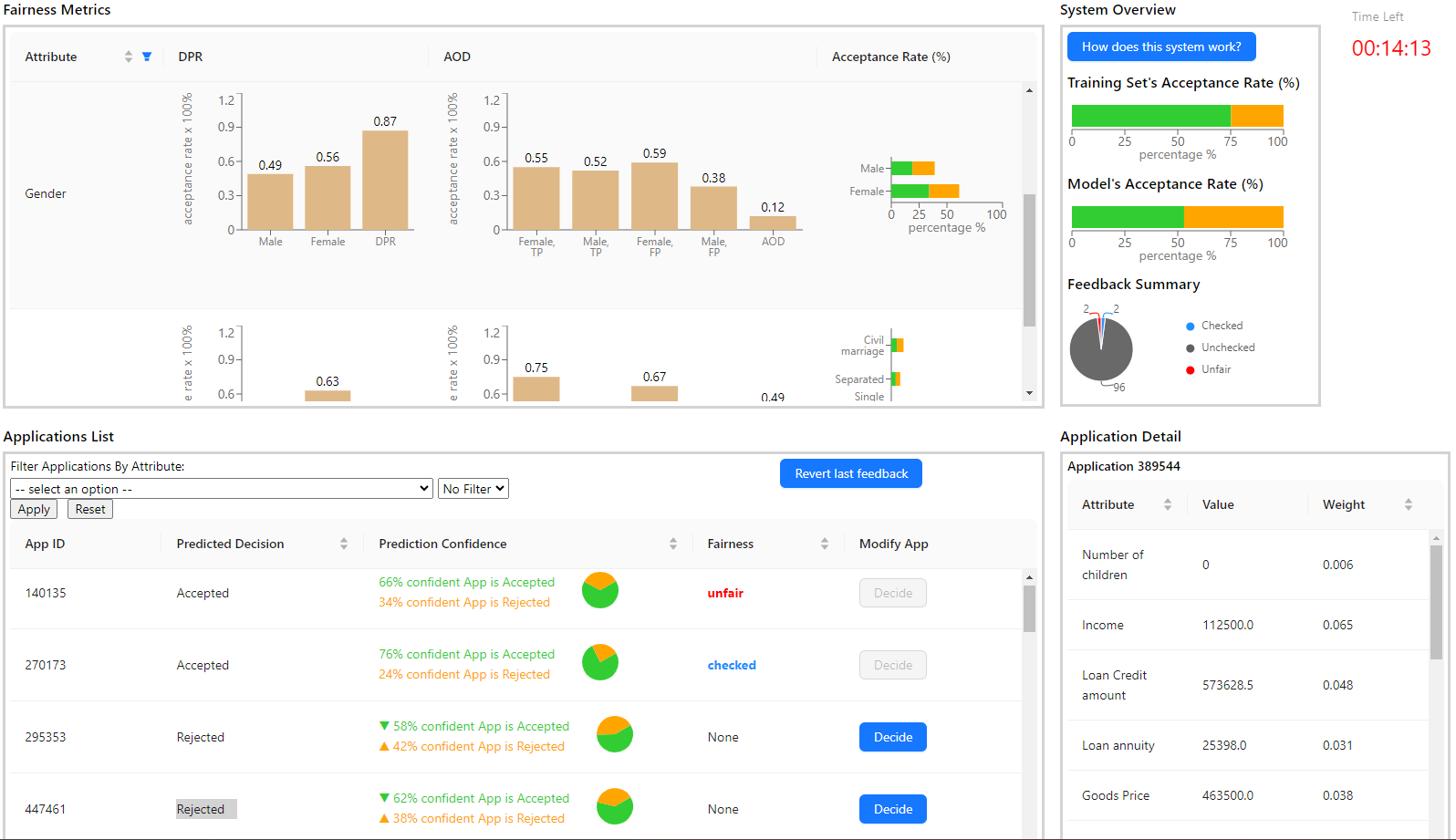}}\\
\subfloat[]{%
    \includegraphics[width=0.4\textwidth]{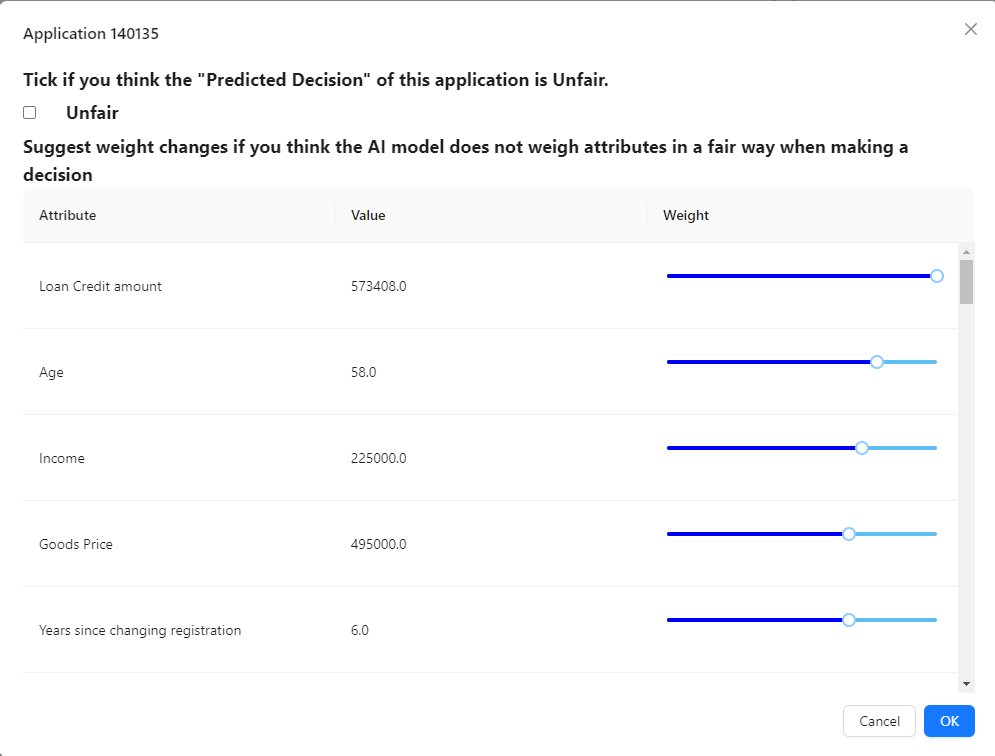}\label{fig:ui_us2b}}%
\hfill\hfill
\caption{(a) The IML UI used in user study 2 to collect participants' feedback on the decisions of the AI model. (b) The modal window shown when participants click the ``Decide'' button of an application appearing on the Applications List. This modal window enables participants to provide their feedback on the fairness of the application's predicted outcome and the ``weight'' the model sets to each attribute in its decision-making.}
\Description{Write 1-2 sentences describing the figure.Used by screen readers to make the document more accessible}
\label{fig:ui_us2}
\end{figure}

Fig.~\ref{fig:ui_us2} presents the UI which we redesigned based on user study 1.  We added additional information to the UI to present fairness metrics changes. To reduce participants' cognitive load and simplify the UI, we only showed DPR and AOD, dropped the model view and the similarity graph and gave them the option to view the tutorial again at any point during the task. The main changes are to the following components:  %and used in our user study to present the details of loan applications, explain how the AI model makes decisions, and enable users to provide and receive feedback. For the design of this UI we followed existing work presenting similar UIs in IML \cite{RN177,RN116,Kulesza2015Principles} and AI fairness \cite{wexler2019if, ahn_fairsight:_2019, cheng2021soliciting, RN54, taka2023,nakao2022}. The prototype is split in four distinct components: the fairness metrics table, the system overview, the applications list table, and the applications detail table. 

\textbf{Fairness Metrics:}
%A major difference to previous work \cite{taka2023,nakao2022} is that we added a fairness metrics component to the UI to explicitly convey quantified information about the fairness of the AI model. Furthermore, we also provide users with information about the impact the feedback they provided had on the fairness of the model, by automatically updating the fairness metrics table, which is a unique feature not made available in similar previous work so far.
This component shows graphs to show the DPR, AOD, and value distribution of the attribute in training set. We show these graphs for Gender, Age and Marital Status by default as they are protected demographic attributes by law and therefore important. The participants could include other attributes and show their details by clicking on the filter button on the header of the Attribute column. The DPR graph shows bars representing the lowest and highest selection rates among all of the attribute values, as well as the DPR value. %the selection rate for the attribute value that is the lowest among all attribute values; the selection rate for the attribute value that is the highest among all attribute values; the DPR of these two selection rates. 
The graph for AOD shows bars representing the lowest and the highest TPR and FPR among all of the attribute values, along with the AOD. %: the highest TPR among all attribute values; the lowest TPR among all attribute values; the highest FPR among all attribute values; the lowest FPR among all attribute values; the average of the differences between the TPs and the FPs (AOD). 
The value distribution graph shows bars for all attribute values and the percentage of them which were predicted to be accepted or rejected. %This is an intuitive measure of fairness.
When users provided feedback on the fairness of the model, the fairness metrics table was automatically updated.

\textbf{Applications List:}
% This component shows a list of $100$ loan applications (i.e., the test set). This list can be filtered by an attribute and a value. The list shows the application ID, the predicted decision by the AI, its confidence in making that decision (the bigger the green ``slice'' of the pie chart is, the more certain the model is), and the assessment the user made in the application details about whether this application decision is unfair or whether they have checked it. The user can sort the applications based on the predicted decision, prediction confidence or their fairness assessment of the applications. Through the UI the user can give feedback to the AI system (through the ``Modify App'' column) to improve its fairness and the AI system can provide feedback about the impact of participant's feedback. 
Clicking on the ``Decide'' button opens a modal window (Fig.~\ref{fig:ui_us2b}). We slightly changed the way that users give feedback: they can only label an application as ``unfair'', and once they give feedback, the application was ``locked'' to prevent multiple feedback instances for the same application. They can adjust the ``weight'' the model assigns specific attributes in its decision-making by dragging the corresponding slider. We showed them the updated weights each time they clicked an application. Clicking ``Cancel'' discards their feedback, leaving the application unchecked and available for assessment. Clicking ``OK'' updates the fairness assessment in the Fairness column to either Unfair or Checked. The model and information displayed in the UI with respect to Predicted Decision and the Predicted Confidence for the unchecked applications, the values of fairness metrics and the acceptance rates of the value distributions and the model (test set) are updated. Importantly, participants can undo their last feedback if they think the impact on the retrained model are undesirable.

\subsubsection{Participants.} We recruited 66 participants aged over 18 years through Prolific who were compensated with £10 per hour; on average the study lasted 35 minutes. We targeted lay users without AI experience but for the purposes of the experiment we did not screen out more experienced users. The study was approved by University of Glasgow College of Science and Engineering Ethics Committee.

\subsubsection{Procedure.} 
After accepting the advertised task on Prolific participants were directed to the study website. After providing informed consent, participants were presented with a pre-task questionnaire about their background, followed by a tutorial to familiarize them with the UI. Subsequently, participants were directed to the main UI and given 20 minutes to provide feedback. After that time was up, they were asked to answer a post-task questionnaire about their experience with using the UI and providing feedback.   

\subsubsection{Data Collected and Analysis.} 
Prolific provides participants' demographic information, which includes age, country of birth, current country of residence, employment status, first language, nationality, sex/gender, and student status. 
Participants recorded whether they had any programming experience or familiarity with machine learning, AI, or statistics in the pre-task questionnaire, and how they would judge fairness in a loan application context.  
In the post-task questionnaire, participants were asked to assess to what extent they trusted AI \cite{hoffman2021} and how much their feedback affected the fairness of the system, what information they used to check the fairness of the system or a specific decision, rate their experience with the UI through the NASA TLX questionnaire \cite{hart1988} and and describe their thoughts about the system.
While they were interacting with the UI, we captured time-stamped logs of users’ actions, e.g., when they included a fairness metric attribute, filtered or sorted the list of applications, clicked on an application, and provided feedback. We also recorded the results of retraining the model. We calculated simple descriptive statistics of our participant profile but we did not conduct any further analysis based on the participants' demographic details. We analysed the subjective feedback of our participants in the pre- and post-task questionnaires by calculating descriptive statistics of their responses, and conducted a thematic analysis of their open responses. We also calculated descriptive statistics of user interactions with the system, based on logging data. We explored the effect of integrating user feedback into a global model and personalized models following the \texttt{Labels\_Unfair} approach and the \texttt{Labels\_Unfair+Weights} approach outlined in section 3.1.6 and 4.1.1.

\subsection{Results}
\subsubsection{Baseline}
Analysing the participants' responses to the pre-task questionnaire, we found that 31 out of 66 participants selected ``People should be treated the same whether they are protected by law or not'', aligning with \emph{counterfactual fairness} \cite{kusner2017counterfactual}. Thus, in addition to the group fairness metrics we already provided in the UI, we decided to calculate the Counterfactual (CF) metric to investigate effects of integrating feedback. We focused on the individualized aspects of counterfactual fairness \cite{maughan2022prediction,black2020} and followed an approach of testing whether the prediction of the deployed model for a certain instance changes when the protected attribute takes a different value \cite{agarwal2018automated,Galhotra2017,black2020} (see the supplemental material for the algorithm we used). The second most selected notion, chosen by 25 participants, was ``Different groups of people who are protected by law should not be disproportionately disadvantaged'', corresponding to group fairness, such as DPR and AOD. Only 9 participants chose ``Similar people will be treated in the same way'', an individual fairness notion.  

Table~\ref{tab:baseline_us2} shows baseline values for Accuracy, Consistency, DPR, AOD, and CF for the three protected attributes. Consistency is relatively high, although no commonly agreed threshold value exists. We found that DPR for Gender is already fair (>0.8) with good values for AOD and CF as well. However, DPR and AOD for Age and Marital Status attributes could be improved, while CF is good for these attributes. Note that these baseline values are higher than in study 1 and cannot be directly compared as we used a different training set. 

\begin{table*}[!h]
\centering
\caption{Baseline metrics for the XGBoost model used in user study 2 with corresponding ``ideal'' values in parentheses. The direction of improvement for a metric is shown in parentheses: $\uparrow$ indicates higher better, $\downarrow$ means lower better. All values are rounded to the second significant digit. }
\label{tab:baseline_us2}
\resizebox{0.45\linewidth}{!}{%
 \begin{tabular}{r|ccc|ccc|ccc|}

    \cline{2-4}
    & \multicolumn{3}{c|}{Baseline in User Study 2} &\multicolumn{3}{c}{}\\
    \cline{2-4}
    
    & \textbf{Acc. $\%$} ($\approxeq 1$) ($\uparrow$) & \textbf{Cons.} ($\approxeq 1$, $\uparrow$) & &\multicolumn{3}{|c}{}\\
     & 69.9 & 0.67 & &\multicolumn{3}{|c}{}\\
      & \textbf{DPR} ($\approxeq 1$, $\uparrow$) & \textbf{AOD} ($\approxeq 0$, $\downarrow$) & \textbf{CF} ($\approxeq 1$, $\uparrow$) &\multicolumn{3}{|c}{}\\
    \textbf{Gender} & 0.85 & 0.14 & 0.88 &\multicolumn{3}{|c}{}\\
    \textbf{Marital Status} & 0.71 & 0.24 & 0.92 &\multicolumn{3}{|c}{}\\
    \textbf{Age} & 0.68 & 0.34 & 0.85 &\multicolumn{3}{|c}{}\\
    
    \cline{2-4}
\end{tabular}%
}
\end{table*}

\subsubsection{Effects of Integrating Feedback on \texttt{Global} Model Fairness}
Table~\ref{tab:global_us2} provides an overview of the effects of integrating all participants’ feedback on the Accuracy and individual and group fairness for the \texttt{Labels\_Unfair} and \texttt{Labels\_Unfair+Weights} feedback integration approaches. We found that Accuracy and Consistency decreased in both feedback integration approaches, similar to study 1 results.

\begin{table*}[!h]
\centering
\caption{Metrics for the retrained \texttt{Global} Xgboost model after integrating the feedback instances from user study 2. Percentage changes $\%$ from baseline are given in parentheses. Improvements against the baseline are highlighted with a shaded background, light gray =<5\%, dark grey >5\%.}
\label{tab:global_us2}
\resizebox{0.6\linewidth}{!}{%
 \begin{tabular}{r|ccc|ccc|}
   
    \cline{2-7}
    
     & \multicolumn{3}{c|}{\texttt{Global\_Labels\_Unfair}} & \multicolumn{3}{c|}{\texttt{Global\_Labels\_Unfair+Weights}}\\
    
    \cline{2-7}
        
      & \textbf{Acc. $\%$} ($\uparrow$) & \textbf{Cons.} ($\approxeq 1$, $\uparrow$) & &
       \textbf{Acc. $\%$} ($\uparrow$) & \textbf{Cons.} ($\approxeq 1$, $\uparrow$)  &\\
     & \makecell[c]{69.2\\(-1.0)} & \makecell[c]{0.66\\(-0.57)}  & & \makecell[c]{68.0\\(-2.72)} & \makecell[c]{0.66\\(-0.82)} &\\      
      
      & \textbf{DPR} ($\uparrow$) & \textbf{AOD} ($\downarrow$) & \textbf{CF} ($\approxeq 1$, $\uparrow$) & \textbf{DPR} ($\uparrow$) & \textbf{AOD} ($\downarrow$) & \textbf{CF} ($\approxeq 1$, $\uparrow$)\\      
     
    \textbf{Gender} &  \makecell[c]{0.83\\(-2)} &  \makecell[c]{0.16\\(16.75)} & \cellcolor{gray!25}\makecell[c]{0.89\\(0.91)} &  \makecell[c]{0.78\\(-7.94)} &  \makecell[c]{0.20\\(45.91)} &  \makecell[c]{0.88\\(-0.23)}\\    
    
    \textbf{Marit. Stat.} &  \cellcolor{gray!25}\makecell[c]{0.75\\(4.69)} & \cellcolor{gray!25}\makecell[c]{0.23\\(-4.53)} & \cellcolor{gray!25}\makecell[c]{0.93\\(0.79)} & \makecell[c]{0.69\\(-2.84)} & 
    \cellcolor{gray!25}\makecell[c]{0.24\\(-2.02)} & \cellcolor{gray!25}\makecell[c]{0.94\\(1.82)}\\
    
    \textbf{Age} & \makecell[c]{0.63\\(-7.50)} &  \makecell[c]{0.39\\(13.27)} & \makecell[c]{0.84\\(-1.20)} & \makecell[c]{0.53\\(-22.92)} & \makecell[c]{0.42\\(23.80)} & \cellcolor{gray!25}\makecell[c]{0.85\\(0.22)}\\ 
     \cline{2-7}%
\end{tabular}%
}
\end{table*}

Surprisingly, nearly all metrics for Gender deteriorated across all approaches, with the notable exception of CF in the \texttt{Labels\_Unfair} approach. However, the changes are so slight that these results can still be considered fair, e.g. DPR on Gender remains >0.8. Instead, participants seemed to focus on the Marital Status attribute, which showed an average improvement of 2.77\% across both integration approaches. Our results also show that fairness metrics for Age mainly deteriorated across approaches, indicating that perhaps participants did not consider it as part of their fairness feedback (we return to this point in section 4.2.3 and 4.2.4). We found that the biggest changes were observed to AOD, similar to study 1, reinforcing its use as an evaluation metric for fairness. We also noted that CF metrics improved but these changes were only small.

\subsubsection{Effects of Integrating Feedback on \texttt{Personalized} Model Fairness.}

\begin{figure}[!h]
\begin{minipage}[b]{\linewidth}
    \centering
 \captionof{table}{Average values of DPR, AOD and CF for the personalized models of all participants in user study 2.}
\label{tab:us2_mean_personalized}
    \resizebox{0.68\textwidth}{!}{%
     \begin{tabularx}{\textwidth}{r|*3{X}|*3{X}|}
    
    \cline{2-7}
    
     & \multicolumn{3}{c|}{\texttt{Personalized\_Labels\_Unfair}} & \multicolumn{3}{c|}{\texttt{Personalized\_Labels\_Unfair+Weights}}\\
     
     \cline{2-7}
     
      & \makecell[c]{\textbf{DPR}($\uparrow$)} & \makecell[c]{\textbf{AOD} ($\downarrow$)} & \makecell[c]{\textbf{CF} ($\uparrow$)} & \makecell[c]{\textbf{DPR}($\uparrow$)} & \makecell[c]{\textbf{AOD} ($\downarrow$)} & \makecell[c]{\textbf{CF} ($\uparrow$)}\\
      
    \textbf{Gender} & \makecell[c]{0.83\\(-1.41)} & \makecell[c]{0.14\\(6.14)} & \makecell[c]{0.87\\(-1.00)} &\makecell[c]{0.84\\(-0.91)} & \makecell[c]{0.14\\(5.22)} & \makecell[c]{0.88\\(-0.73)}\\
    
    \textbf{Marit. Stat.} & \makecell[c]{0.71\\(-0.29)} & \cellcolor{gray!25}\makecell[c]{0.24\\(-0.21)} & \cellcolor{gray!25}\makecell[c]{0.93\\(1.30)} & \cellcolor{gray!25}\makecell[c]{0.71\\(0.04)} & \cellcolor{gray!25}\makecell[c]{0.24\\(-2.76)} & \cellcolor{gray!25}\makecell[c]{0.93\\(1.20)}\\
    
    \textbf{Age} & \makecell[c]{0.64\\(-6.93)} & \cellcolor{gray!50}\makecell[c]{0.32\\(-5.10)} & \cellcolor{gray!25}\makecell[c]{0.87\\(2.18)} & \makecell[c]{0.64\\(-5.58)} & \cellcolor{gray!50}\makecell[c]{0.32\\(-6.80)} & \cellcolor{gray!25}\makecell[c]{0.87\\(2.20)}\\
    \cline{2-7}%
\end{tabularx}%
}
\end{minipage}\\
\begin{minipage}[b]{\linewidth}
\centering
\includegraphics[width=0.6\linewidth]{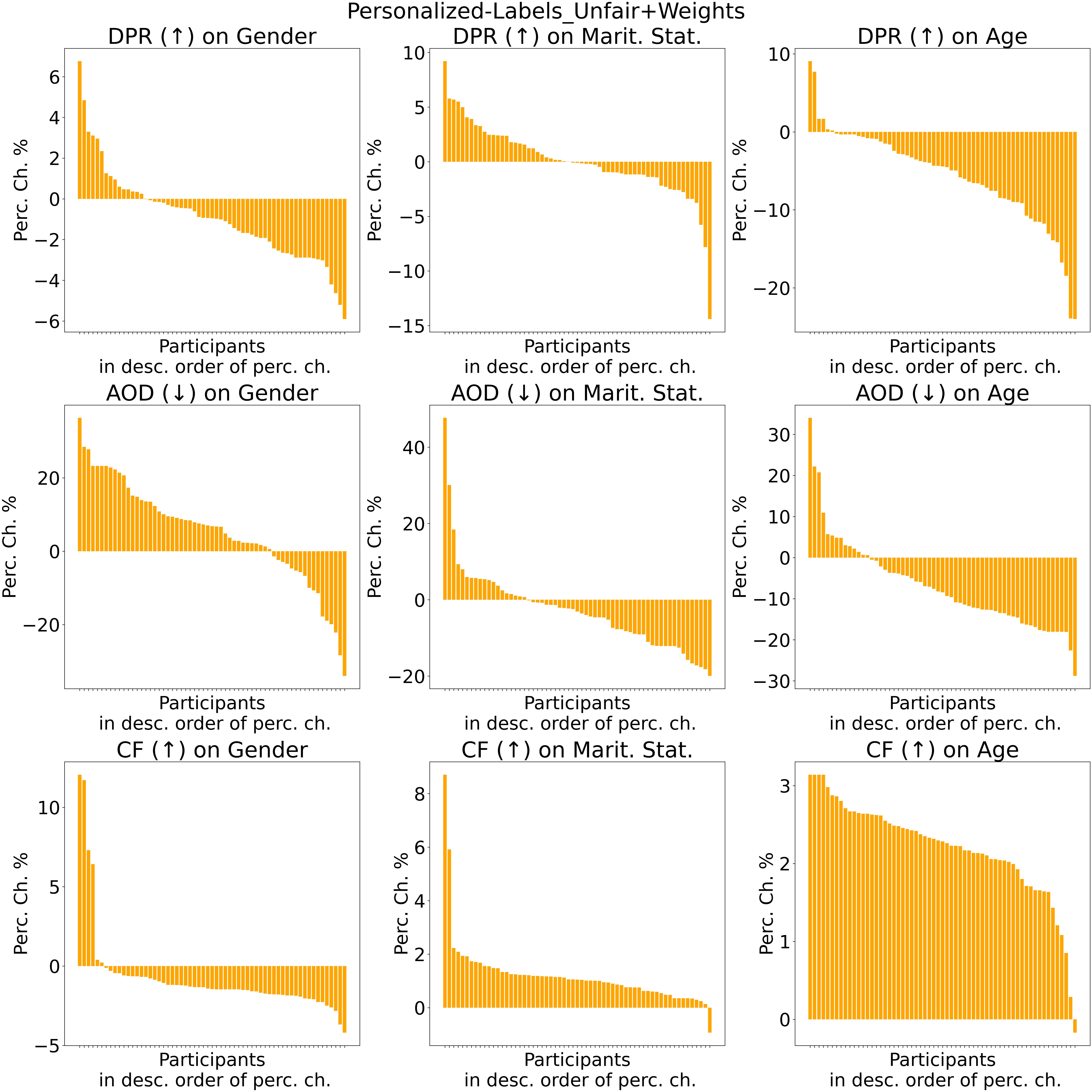}%
\captionof{figure}{Percentage change of DPR, AOD and CF from baseline per participant for the \texttt{Personalized-Labels\_Unfair+Weights} approach, presented in bar graphs. }
\Description{The top line shows 9 bar graphs for the \texttt{Personalized-Labels\_Unfair+Weights} feedback integration method: 3 for DPR, 1 each for gender, marital status and age, 3 for AOD, 1 each for gender, marital status and age, 3 for CF, again 1 each for gender, marital status and age. It shows that for DPR, roughly half the participants improved the metric for marital status, many fewer improved it for gender, while almost everyone deteriorated for age. For AOD, over half the participants improved the metric on age, fewer improved it for marital status and even fewer for gender.}
\label{fig:us2_indiv_models}
\end{minipage}
\end{figure}

Similar to study 1, we then investigated the effects of participants' feedback on their personalized models. Table \ref{tab:us2_mean_personalized} shows the average values of the fairness metrics across all participants. Our results show that, on average, participants deteriorated all of the fairness metrics on Gender but still stayed within a range which could be considered fair, similar to the results for the global models. Further, they improved fairness on Marital Status for most of the fairness metrics across both approaches, again similar to the global model results. However, in their personalized models they also improved AOD and CF on Age, with some large improvements on AOD in particular for the \texttt{Personalized-Labels\_Unfair+Weights} approach.

Investigating the per-participant percentage changes of the DPR, AOD and CF fairness metrics for the approach \texttt{Personalized-Labels\_Unfair+Weights} (Fig.~\ref{fig:us2_indiv_models}) shows the effect of participants' feedback on personalized models in more detail. There are some participants that improved the fairness metrics on Gender but they are outweighed by participants who did not. For Marital Status we note that AOD and CF fairness metrics contain some outliers that affected the average across all participants.  Whereas with the global models we did not see any improvements on Age, we saw that indeed, on average, personalized models resulted in an improvement. Most participants had improved AOD on Age by a large amount, slightly counterbalanced by some which deteriorated the metric by a large amount. For CF on Age, almost all participants improved this metric by a small amount. We will return to the differences between global and personalized models in the Discussion.

\subsubsection{Other Feedback}
Similar to study 1, we were interested in which attributes might matter to users, in addition to protected attributes. We already provided participants with fairness information for three protected attributes by default so we looked at what they changed. We found that only 2 participants removed protected attributes; one excluded \textit{Age} and the other \textit{Marital Status}. Very few participants added other non-protected attributes: the top most frequent were \textit{Number of children, Income, Years in current employment, Highest education level, Years since changing registration, Occupation Type, Employer organization}, and \textit{Owns Property} with no more than 3 participants selecting any of them (find full list of attributes in supplementary material). 

Results of interacting with these attributes was similar to study 1 (see Table~\ref{tab:attributes_interaction} and find full list in supplementary material). We found that the top attributes involved in filtering applications included \textit{Age} (18 participants), \textit{Gender} (11 participants), \textit{Income} (7 participants) and \textit{Marital Status} (6 participants) and \textit{Accompanied while applying} (6 participants). This is almost identical to what we found in study 1. We also investigated the top attributes that participants interacted with. While 23 participants in study 1 interacted with Age and increased the weight of this feature, 44 participants in study 2 fed back that Age should be \textit{less} important in the decision-making, demonstrated by decreasing the weight. Similar to study 1, the top attributes interacted with  include \textit{Owns Car, Income type, Owns Property, Income, Years in current employment}, and \textit{Highest education level}. Other attributes were \textit{Number of children, loan credit amount} and \textit{Years since changing registration}. 

\begin{table}[ht]
    \centering   
    \caption{Top 12 attributes that participants interacted with in at least one of their feedback instances in user study 2, along with the average proposed weight change.}%
    \label{tab:attributes_interaction}
    \resizebox{0.7\columnwidth}{!}{%
    \rowcolors{2}{gray!25}{white}
 \begin{tabular}{rc|rcc}
    \hline
    \multicolumn{2}{c|}{\textbf{Filtering for Attribute}} & \multicolumn{3}{c}{\textbf{Attribute Weight Changes}}\\
    
     \hline
      \textbf{Attribute} & \makecell[r]{\textbf{No.} \textbf{Part.}} &  \textbf{Attribute} & \makecell[r]{\textbf{No.} \textbf{Part.}} & \makecell[r]{\textbf{Av. Weight} \textbf{Ch. $\%$}}\\
   \hline
   
    \textbf{Age} &18  & 
    % Number of children & 3 & 
    \textbf{Age} & 44 & -1.53\\

    \textbf{Gender} & 11 & 
    % Income & 3 & 
    Income & 39 & 7.44\\

    Income & 7 & 
    % \makecell[r]{Years in\\current employment} & 2 & 
    \makecell[r]{Years in current employment} & 33 & 3.04 \\

    \textbf{Marital Status} & 6 &  
    % \makecell[r]{Highest\\education level} & 2 & 
    Owns Car & 28 & 4.66\\

     \makecell[r]{Accompanied while applying} & 6 & 
     % \makecell[r]{Years since\\changing registration} & 2 & 
     Loan Credit amount & 28 & -11.93\\

     Loan Credit amount & 4 & 
     % Occupation Type & 2 & 
     Income type & 28 & 99.54\\

    \makecell[r]{Highest education level} & 3 & 
    % Employer organization & 2 & 
    Owns Property & 27 & 119.96\\

     Number of children & 3 & 
     % Owns Property & 2 & 
     \makecell[r]{Employer organization} & 27 & 20.85\\

     Has Mobile & 2 & 
     % Income type & 1 & 
     \textbf{Gender} & 26 & $\inf$\\

     Employer organization & 1 & 
     % Region \& City Rating & 1 & 
     Goods Price  &  25 & 0.67\\

     Application Day & 1 & 
     % Has work phone & 1 & 
     Occupation Type  &  25 & 17.79\\

     Owns car & 1 & 
     % Owns car & 1 & 
     \textbf{Marital status}  &  23 & 46.93\\
    \hline
\end{tabular}%
}
\end{table}%

\subsubsection{UI Interactions and Usability}
Finally, we investigated how participants used the UI and approached the task. We found that the mean trust score was 3.01 (sd=0.65) on a 7-point Likert scale, indicating that participants neither trusted nor particularly distrusted the AI. Their responses indicated high wariness of the AI's decision-making, which is appropriate for finding problems with fairness.

The UI provided a number of components that supported participants in assessing fairness. When asked how they checked fairness, a majority of participants told us they referred to information made available through the UI. Although participants used formal metrics we provided, such as the DPR (36 responses) and AOD (29 responses), the most frequently reported component used to check fairness was the \textit{value distributions} (39 responses). These graphs provided an intuitive way to assess group fairness and essentially communicate the same information as the DPR but are less daunting. However, a surprisingly large proportion of participants used ad-hoc ways to assess fairness, such as looking through applications and making intuitive judgements (35 responses), or sorting/filtering applications to determine patterns of decisions (27 responses). 

\begin{figure}[!h]
    \centering
    \includegraphics[width=0.8\linewidth]{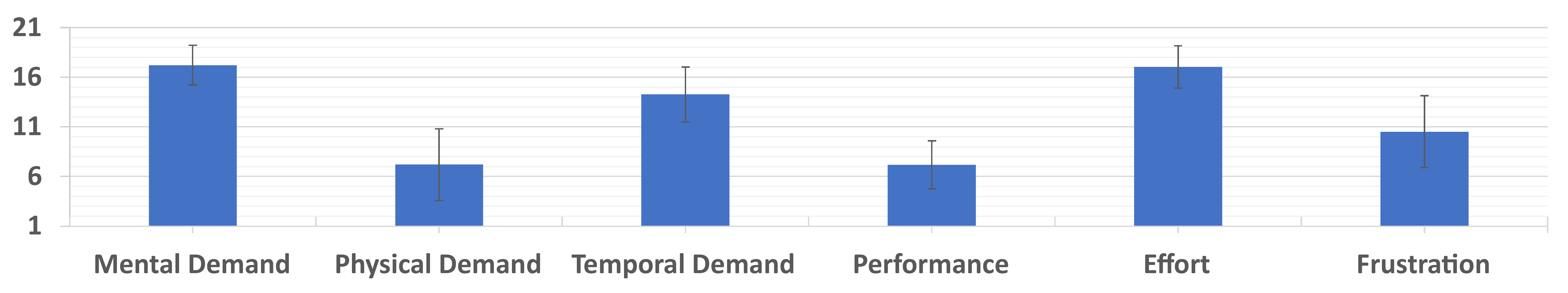}%
    \caption{Average NASA TLX scores.}
    \label{fig:NASATLX}
    \Description{The figure shows 6 error bars to present the NASA TLX scores of participants in user study 2. The bars correspond to Mental Demand, Physical Demand, Temporal Demand, Performance, Effort, and Frustration. The average scores for Mental Demand, Temporal Demand and Effort are quite high, greater than 14.}
\end{figure}

The NASA TLX questionnaire results show that this was indeed a complex task (Fig.~\ref{fig:NASATLX}). We observed that mental demand (m=17.21, sd=3.98), temporal demand (m=14.27, sd=5.53) and effort (m=17.04, sd=4.29) measures were relatively high, while performance (m=7.18, sd=4.85) and physical demand (m=7.19, sd=7.23) were low. These results are substantiated by their qualitative responses which showed that participants found the UI complex and at times confusing (21 responses),  with too much information and not enough time to provide good feedback (17 responses). Some participants also stated that they would like more information (7 responses) but this would need to be carefully balanced so as not to overwhelm users. However, we also received responses from 11 participants that commended our UI as being  easy, simple, and clear. It seems that while our UI can support stakeholders well enough, there is room for improvement. We will return to this point in the Discussion.

\section{Discussion}
\subsection{Limitations}
Our studies have three major limitations. First, there are shortcomings in the \emph{setup of the studies}. We only involved a relatively small sample of people, and we did not stratify them in terms of level of expertise in or knowledge of machine learning. Participants had only 20 minutes to provide feedback and our results showed that many felt this put them under time pressure. This might affect the nature and quality of the feedback that participants gave. We also did not randomize the order of the applications shown to participants which led to many participants providing feedback for the same applications. This issue also affects \emph{models and the feedback} we used. Integrating the participants' feedback on the same applications would not be very informative because it comprises repetitive training instances with possibly conflicting labels.  Further, we encountered some problems with weights and binning of attribute values. When participants changed a weight, the distribution of weights stopped being a valid distribution summing to 1;  thus, we normalized the weights after each feedback but improvements could be made how to elicit weights from users. Similarly, we made some decisions around binning of numerical attribute values which influences the calculation of fairness metrics. Perhaps we could turn to users to suggest acceptable bins. Last, our investigations were constrained to the loan application and credit risk domain, where the models we created were already relatively fair. 

\subsection{Implications for Responsible AI and Human-in-the-loop Fairness}

\textbf{Marking up unfair instances and providing weight feedback seems to work.} It could be argued that users might find it difficult to identify specific decisions that they find fair or unfair, however, previous works using similar approaches have shown success \cite{cheng2021soliciting, nakao2022}. Our work points to the possibility of leveraging user feedback to determine what fairness metric to use, as well as what attributes to include. In study 1, the \texttt{Labels\_Unfair} and  \texttt{Labels+Weights} approaches improved DPR, EOD and AOD; the same approaches worked in study 2 to improve AOD and CF. We also found that participants focused on Marital Status and Age, along with other non-protected attributes. 

\textbf{The approach could be used throughout the AI development lifecycle.} Current AI legislation, such as the European AI Act, call for regular assessment for high-risk AI systems throughout their development. It is still to be determined how these regular assessment are to take place but HITL fairness feedback could be obtained before a model is deployed in practice, to reflect stakeholder views more closely. Although our approach gathered user feedback after an initial model was trained, user feedback could also be gathered on the training set, using labels to mark Fair and Unfair decisions. Finally, it could even be useful as a testing approach after deployment, to ensure that models stay fair.

\textbf{Fairness is multi-faceted and complex.} When participants labelled a prediction as unfair, were they assessing fairness or the ``correctness'' of decision-making? We believe that these distinctions are not really helpful as they only serve to prioritize outcome fairness over procedural fairness. Instead, our results corroborate previous research \cite{jakesch_how_2022} that people hold conflicting fairness notions in the same context.  Evidence for complex fairness notions in our studies comes through the interaction with non-protected attributes: we found that less than half of the participants interacted with the protected attributes, while there were other, non-protected attributes participants might have deemed as irrelevant or unfairly used (e.g. \texttt{Owns Car}) or more important (e.g. \texttt{Years in current employment}). Perhaps these participants considered other biases within the dataset and the subjective nature of ground truth in loan applications. Thus, user feedback is vital to extend fairness metrics beyond protected attributes to reflect the perspectives of stakeholders.  

\textbf{Caution is advised on choosing fairness metrics to assess.} Using user feedback means we need to be careful when measuring fairness with specific metrics, especially if users do not get any information how their feedback affects these metrics. In study 2, we based our choice of fairness metrics to display on results of study 1 which showed that DPR, EOD, and AOD were improved in the ``global'' results (Table~\ref{tab:Global_results_us1_all}). However, when we asked what fairness notions participants were subscribing to, many responded with CF. Thus, we need interactive ways of showing a range of fairness metrics for people to choose from.

\subsection{Future Work}
\textbf{Finding better ways to support feedback.} There are many ways in which better feedback could be obtained, in addition to what we have already discussed in Limitations. Recall that the UI was not designed to provide information about Consistency or Counterfactual fairness. More information that could help in assessing fairness could be embedded in the UI but we need to be mindful of the complexity of the task, risking information overload. Further, individual fairness could be measured in other ways, such as the Theil index \cite{speicher2018unified}. We could also allow users to provide direct feedback on which fairness metrics and attributes they are trying to optimize. Fairness metrics also need to be better explained but again this risks adding to the complexity of the UI. Certainly, the process of assessing fairness requires an unrushed task setting and we suggest extending feedback time to at least one hour.

\textbf{Developing new ways of measuring fairness.} Recall that participants interacted with non-protected attributes such as \texttt{Income}, where common outcome fairness metrics might be inappropriate to apply. As a community interested in AI fairness, we need to decide standard ways of measuring fairness for comparison. 

\textbf{Investigating other domains and stakeholders.} While our focus was on feedback by lay users, other stakeholders may have other fairness definitions, driven possibly by their job responsibilities~\cite{Nakao2023Stakeholder}; thus, research that includes a wider range of stakeholders and their fairness notions is warranted. Similarly, we only worked with one dataset, mainly because of the context-dependent nature of fairness. Hence, it would be interesting to investigate how this approach can extend to other datasets and domains, as well as investigate cultural aspects of fairness \cite{nakao2022}.

\textbf{Extending ways to integrate user feedback.} We have only scratched the surface of different ways to integrate user feedback. Currently, our global approaches of integrating feedback simply aggregates feedback and thus “blends” fairness perceptions. We already noted that there are many repeated training instances in the global models and thus the effects in personalized models are much stronger. Thus, future research could address what to do with conflicting fairness perceptions, e.g. how to best integrate personalized models and feedback into a global model. 

\textbf{Addressing possible gaming of feedback and ``bad actors''.} A bad actor could actually inject \emph{more} unfairness into a model. As previous research \cite{nakao2022} and our own results have found, some participants gave “poor” feedback and therefore deteriorated fairness, at least when measured on the fairness metrics and attributes we implemented. There could be a number of different approaches to deal with this. First, we need to identify what bad feedback consists of. This is problematic as it depends on the choice of fairness metric. Second, we would need to identify the actors who consistently give bad feedback. Last, we need to develop approaches how to deal with these actors and feedback. For example, we could choose to ignore these actors completely, and discard their personalized models. Another option would be to simply not integrate particular feedback.

\section{Conclusion}

In this work, we explored methods to integrate users’ feedback on AI fairness and investigated impacts on fairness metrics. We ran two user studies to collect feedback from participants on the fairness of predicted labels from a pre-trained XGBoost model for loan applications, allowing the adjustment of feature weights. In study 1, we obtained feedback from 58 participants, and investigated three different feedback integration approaches in global and personalized models. In study 2, 66 lay users provided instance-level feedback in an IML setting and we investigated two integration approaches. Overall, we found that: 
\begin{itemize}
    \item Allowing users to mark up Unfair labels and weight changes to attributes provides a good approach to reflect stakeholders' AI fairness perspectives, even if they have no technical background.
    \item User feedback can improve fairness metrics, especially AOD and CF for protected attributes. 
    \item Participants focused on a range of features including non-protected attributes. Many participants adjusted weights of non-protected attributes, indicating that they might have not used common fairness metrics to judge AI fairness. We argue that these attributes and new fairness metrics need to be considered in AI fairness. 
    \item Participants were able to use the UI but it could be improved. Participants frequently felt overwhelmed and under time pressure.  Fairness assessments require a lot of information to be communicated and cannot be rushed, to counteract users relying on ad-hoc ways or intuition.
\end{itemize}

Overall, our findings demonstrated the feasibility of incorporating stakeholders’ feedback for AI fairness but also pointed out its challenges. By making the data and analysis code available to bootstrap further comparison and research, we have taken important first steps toward leveraging user feedback in the design and development of fair AI. We encourage other researchers and practitioners to build on our foundational work and develop new approaches for integrating stakeholder feedback so that organisations can start to adopt more participatory approaches as part of responsible AI development.

\begin{acks}

We thank Rajat Choudhary for model development and Yiyu Zhong for UI implementation for user study 1.
\end{acks}

\bibliographystyle{ACM-Reference-Format}
\bibliography{manuscript_base}

\end{document}